\documentclass{article}

\usepackage{graphicx}
\usepackage{subcaption}
\usepackage{booktabs}

\PassOptionsToPackage{table}{xcolor}

\usepackage{hyperref}

\usepackage[preprint]{icml2026}

\usepackage{amsmath}
\usepackage{amsfonts}
\usepackage{amssymb}
\usepackage{mathtools}

\usepackage[protrusion=false,expansion=false]{microtype}

\usepackage{longtable}
\usepackage{placeins}
\usepackage{csvsimple}
\usepackage{array}
\usepackage{threeparttable} 
\usepackage{adjustbox}
\usepackage{multirow}
\usepackage{makecell}
\usepackage{float}
\usepackage{enumitem}

\usepackage{tikz}  
\usetikzlibrary{shapes.geometric, arrows, positioning, fit}
\usepackage{pgfplots}
\usepgfplotslibrary{groupplots}
\pgfplotsset{compat=1.18}



\DeclareMathOperator{\sg}{sg}

\renewcommand{\algorithmicrequire}{\textbf{Input:}}
\renewcommand{\algorithmicensure}{\textbf{Output:}}
\newcommand{\RETURN}{\STATE \textbf{return} }

\icmltitlerunning{VEPO: Variable Entropy Policy Optimization}

\begin{document}

\twocolumn[
  \icmltitle{VEPO: Variable Entropy Policy Optimization for \texorpdfstring{\\}{ }
    Low-Resource Language Foundation Models}


\begin{icmlauthorlist}
  \icmlauthor{Chonghan Liu}{qy}
  \icmlauthor{Yimin Du}{qy}
  \icmlauthor{Qi An}{qy}
  \icmlauthor{Xin He}{qy}
  \icmlauthor{Cunqi Zhai}{qy}
  \icmlauthor{Fei Tan}{qy}
  \icmlauthor{Weijia Lin}{qy}
  \icmlauthor{Xiaochun Gong}{qy}
  \icmlauthor{Yongchao Deng}{qy}
  \icmlauthor{Shousheng Jia}{qy}
  \icmlauthor{Xiangzheng Zhang}{qy}
\end{icmlauthorlist}

\icmlaffiliation{qy}{Qiyuan Tech}
\icmlcorrespondingauthor{Chonghan Liu}{khazzz1c@gmail.com}

  \icmlkeywords{Machine Learning, ICML, Machine Translation, Low-Resource Languages, Reinforcement Learning, VEPO}

  \vskip 0.3in
]


\printAffiliationsAndNotice{}
\begin{abstract}
    Large language models frequently exhibit suboptimal performance on low resource languages, primarily due to inefficient subword segmentation and systemic training data imbalances. In this paper, we propose \textbf{Variable Entropy Policy Optimization (VEPO)}, which leverages Reinforcement Learning with Verifiable Rewards to incorporate deterministic structural constraints into the policy alignment process. This framework ensures prescribed sequence length, robust format consistency, and rigorous linguistic well formedness, all enforced during training. Central to our approach is a variable entropy mechanism that enables the model to dynamically calibrate the equilibrium between literal fidelity and semantic naturalness by modulating the exploration exploitation manifold. By integrating entropy tempered advantage estimation with asymmetric clipping, VEPO sustains robust exploration while mitigating policy collapse. Empirical evaluations across 90 FLORES-200, COMET-22, chrF directions demonstrate that VEPO yields substantial improvements in both tokenization efficiency and translation quality, bridging the performance gap for underrepresented languages.
\end{abstract}

\section{Introduction}
\label{sec:intro}
While Large Language Models have revolutionized Neural Machine Translation, their performance in low-resource regimes remains hampered by suboptimal tokenization, training imbalances, and reinforcement learning instabilities. State-of-the-art models such as GPT4, DeepSeek-R1 and Qwen-max\citep{openai2024gpt4technicalreport,
Guo_2025,qwen2025qwen25technicalreport} frequently exhibit pronounced sequence fragmentation and substantial variance in subword segmentation for morphologically rich writing systems, such as Khmer and Thai, predominantly as a consequence of vocabularies and training corpora that are disproportionately optimized for high-resource languages. Specialized MT architectures \citep{cheng2025seedxbuildingstrongmultilingual, zheng2025hunyuanmttechnicalreport, dou2025sailor2sailingsoutheastasia} address these gaps via curated data, yet often struggle with the instruction flexibility required for production environments.

\begin{figure}[t]
    \centering
    \includegraphics[width=\columnwidth]{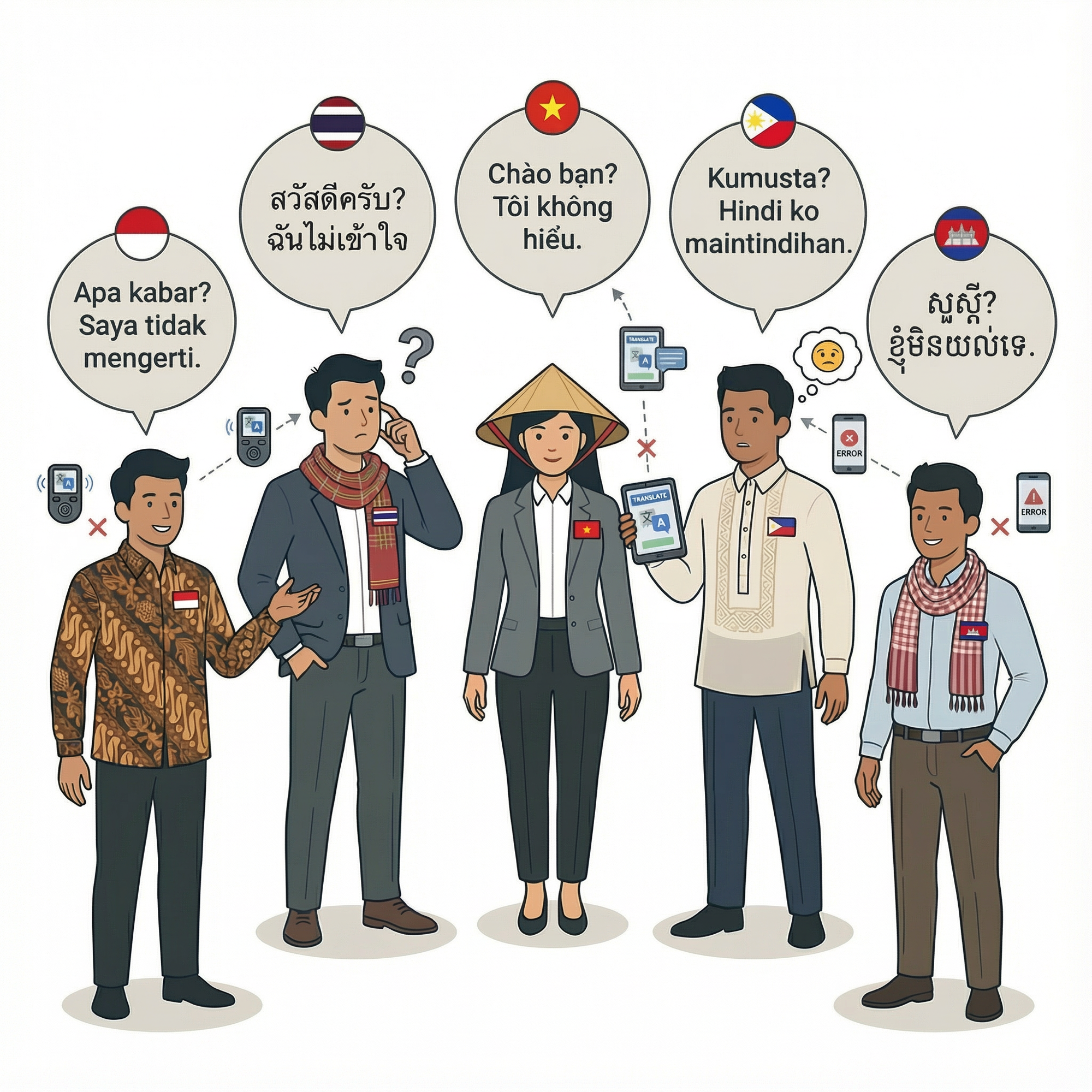}
    \label{fig:main_results}
\end{figure}

We identify three primary obstacles to low-resource NMT: (1) \textbf{Tokenization Inefficiency}, in which conventional subword segmentation methods induce a 3-4$\times$ increase in sequence fragmentation; (2) \textbf{Linguistic Marginalization} due to English-centric data distributions; (3) \textbf{Policy Collapse}, characterized by entropy decay and verbosity in standard RL. To mitigate these, we propose Variable Entropy Policy Optimization, VEPO is further augmented by Reinforcement Learning with Verifiable Rewards, which enforces deterministic structural constraints directly during optimization—addressing training-time failures that test-time scaling \citep{bajpai2025multilingualtesttimescalinginitial} cannot resolve.

\noindent\textbf{Contributions.} We deliver: \textbf{(i)} a tokenizer-driven continued pre-training strategy that significantly enhances subword efficiency for low-resource scripts; \textbf{(ii)} a variable entropy mechanism that enables dynamic calibration between literal fidelity and semantic naturalness; and \textbf{(iii)} an RLVR-integrated framework that effectively mitigates pathologies such as length redundancy and linguistic drift. Extensive evaluations across 90 FLORES-200 directions demonstrate that our model achieves state-of-the-art performance across \textbf{BLEU}, \textbf{COMET} and \textbf{chrF} metrics, while maintaining competitive zero-shot accuracy on general reasoning and instruction-following benchmarks (Table~\ref{tab:translation_results}).
\section{Related Work}
\label{sec:Related work}

Modern multilingual LLMs endeavor to harmonize high-fidelity translation with general purpose instruction following, typically through architectural specialization or sophisticated objective blending. \textit{Apertus} is developed with a primary focus on methodological transparency and extensive multilingual coverage—particularly for low-resource and otherwise underrepresented languages—while intentionally avoiding reliance on explicit, task-specific translation templates \citep{apertus2025apertusdemocratizingopencompliant}. \textit{TOWER+} exemplifies a phased adaptation regime, integrating continued pre-training, supervised fine-tuning, and preference optimization with verifiable rewards to balance translation accuracy with conversational utility \citep{rei2025towerbridginggeneralitytranslation}. Concurrently, \textit{EMMA-500} and \textit{LLaMAX} investigate strategies for massively multilingual adaptation \citep{ji2025emma500enhancingmassivelymultilingual,lu2024llamaxscalinglinguistichorizons}, while \textit{Aya} explores the efficacy of instruction tuning in diverse linguistic contexts \citep{ustun2024ayamodelinstructionfinetuned}.

\begin{figure}[h]
    \centering
    \includegraphics[width=\columnwidth]{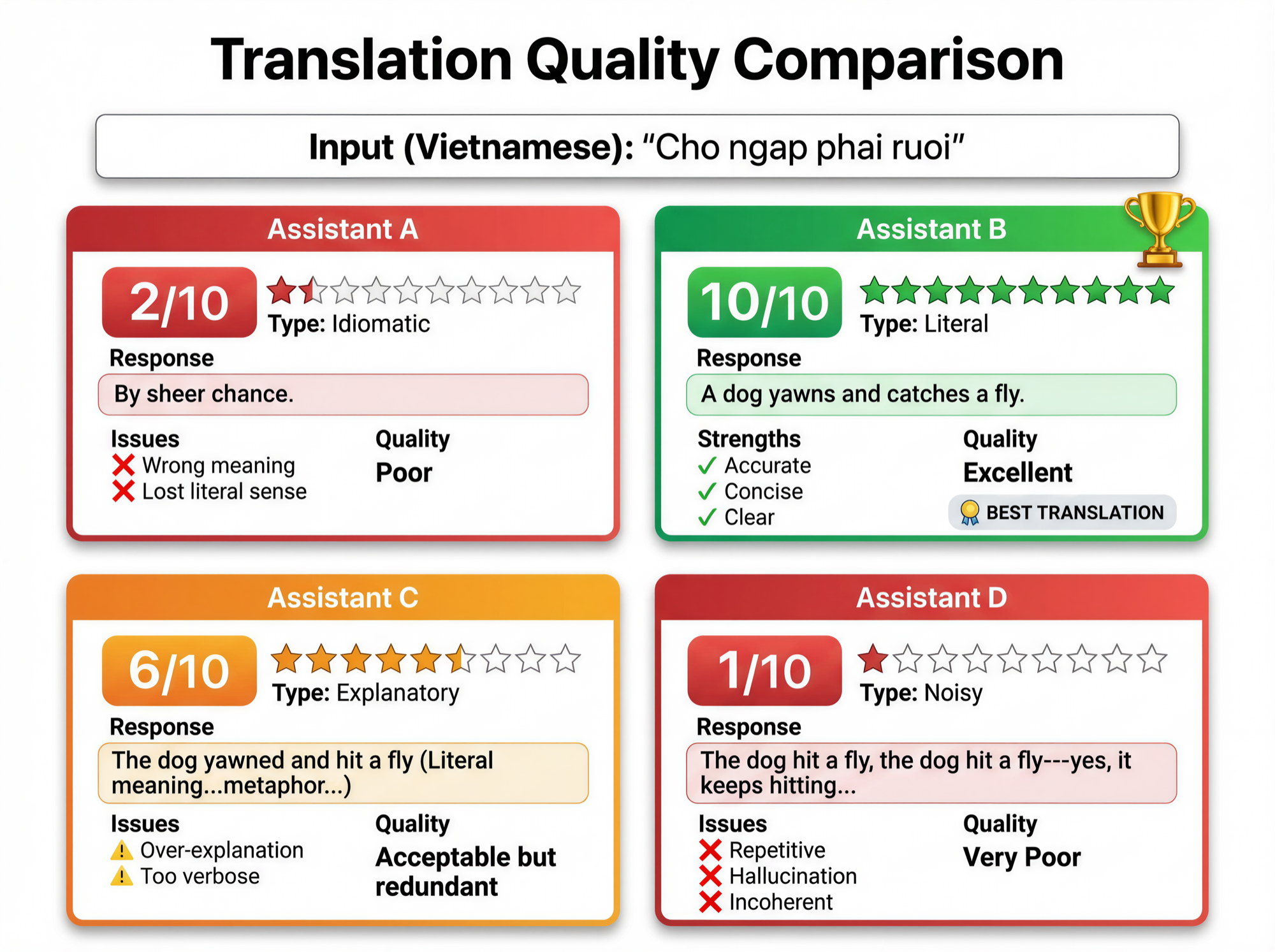}
    \caption{Comparison of translation quality assessment across different models.}
    \label{fig:translation_radar}
\end{figure}

\noindent\textbf{Regional Models for ASEAN Languages.} Several initiatives focus on Southeast Asian (SEA) linguistic landscapes through curated data and targeted alignment. \textit{COMPASS-V2} targets SEA e-commerce scenarios by optimizing the synergy between curated corpora and bespoke tokenization \citep{maria2025compassv2technicalreport}. \textit{SEA-LION} \citep{ng2025sealionsoutheastasianlanguages}, \textit{Sailor2} \citep{dou2025sailor2sailingsoutheastasia}, and \textit{SeaLLMs 3} \citep{zhang2024seallms3openfoundation} extend regional coverage through continual pre-training and culture-aware sampling. Furthermore, \textit{Babel} \citep{zhao2025babelopenmultilinguallarge} demonstrates how expansive multilingual scaling can effectively complement localized regional efforts.

\noindent\textbf{Dedicated Machine Translation Systems.} Specialized NMT models advance the state-of-the-art by prioritizing bilingual fidelity through task-specific architectures. Seed-X \citep{cheng2025seedxbuildingstrongmultilingual}, Qwen-MT \citep{qwenmt2025wherespeedmeetssmarttranslation}, and Hunyuan-MT \citep{zheng2025hunyuanmttechnicalreport} demonstrate that translation-centric designs can achieve performance parity with commercial systems. Recent compact models, such as X-ALMA \citep{xu2025xalmaplugplay}, further optimize efficiency via modular adapters and selective rejection mechanisms, elucidating optimization levers that remain potent at smaller parameter scales.

\noindent\textbf{Reinforcement Learning for NMT.} Policy optimization in NMT has evolved from minimum risk training, which directly optimizes sequence-level metrics \citep{shen-etal-2016-minimum}, to sophisticated preference-driven alignment. Recent work on Contrastive Preference Optimization (CPO) sharpens reward landscapes for translation tasks \citep{xu2024contrastivepreferenceoptimizationpushing}, while tailored RLHF pipelines address cost-sensitive preference modeling \citep{xu2024advancingtranslationpreferencemodeling, zhu2024preferencedrivenparadigmenhancedtranslation}. Critically, diagnostic studies have elucidated the fundamental weaknesses of unconstrained RL objectives and evaluation protocols \citep{choshen2020weaknessesreinforcementlearningneural}, prompting recent extensions into multilingual preference optimization \citep{dang2024rlhfspeaklanguagesunlocking}.

\noindent\textbf{Limitations vs.\ VEPO.} Despite their versatility, general-purpose LLMs often exhibit three systemic failure modes in translation: (1) \textit{Fidelity Gaps}, where the absence of task-specific constraints leads to semantic hallucinations or overt translation errors; (2) \textit{Verbosity Bias}, characterized by redundant supplementary explanations or conversational filler that detracts from concise output (see Figure~\ref{fig:translation_radar}); and (3) \textit{Generation Overrun}, a phenomenon where the model continues generating irrelevant text or "hallucinatory continuations" after the target translation is completed. VEPO diverges from existing multilingual paradigms by addressing these issues across three dimensions. First, unlike specialized architectures such as Qwen-MT \citep{qwenmt2025wherespeedmeetssmarttranslation} that are often confined to rigid templates, VEPO maintains expansive instruction-following capabilities while ensuring bilingual accuracy. Second, VEPO mitigates redundancy and over-generation through the synergy of RLVR-integrated structural constraints and length-invariant reinforcement learning normalization, providing robust control over sequence termination. Third, by introducing a variable entropy mechanism, VEPO enables the model to dynamically calibrate the trade-off between literal fidelity and semantic naturalness, facilitating both exact and paraphrastic translations beyond the limitations of traditional alignment objectives.

\section{Methods}
\label{sec:methods}
To ensure a robust foundation for effective reinforcement learning on low-resource corpora, we expand the tokenizer to mitigate subword fragmentation, followed by continued pre-training on billion-scale multilingual corpora to establish a strong linguistic baseline. During the supervised fine-tuning phase, we supplement the model with high-quality bilingual translation data and instruction-following datasets covering both high-resource and underrepresented languages. Finally, we apply Variable Entropy Policy Optimization to execute entropy-aware RL, achieving precise policy alignment while maintaining stylistic flexibility.

\subsection*{Continued Pre-training}
Continued pre-training serves to adapt large language models to specialized linguistic domains while preserving general-purpose capabilities \citep{conneau-etal-2020-unsupervised}. In low-resource scenarios, the architectural design of the vocabulary is paramount. Introducing language-specific tokens facilitates the capture of distinct morphological structures \citep{conneau2020unsupervisedcrosslingualrepresentationlearning,fan2020englishcentricmultilingualmachinetranslation,devlin-etal-2019-bert}, thereby mitigating tokenization inefficiency through optimized subword segmentation \citep{sennrich-etal-2016-neural,liu-etal-2018-visual}. 

\textbf{Stage 1: Vocabulary Expansion.} We extend the Qwen2.5-7B tokenizer \citep{qwen2025qwen25technicalreport} by integrating dedicated tokens for eight low-resource languages, yielding the Qwen2.5-7B-8Langs tokenizer. This expansion follows established multilingual methodologies \citep{nllbteam2022languageleftbehindscaling} to ensure robust coverage. 

\textbf{Stage 2: Balanced Multilingual Training.} To prevent catastrophic forgetting and maintain English proficiency, we employ a 1:1 sampling ratio between English and low-resource corpora. The expanded vocabulary introduces approximately 3k--4k tokens per target language, achieving a $2.1\times$--$5.4\times$ improvement in compression ratios. This efficiency gain translates directly into reduced inference latency and enhanced computational throughput.

\subsection*{Post-Training Alignment}

\paragraph{Supervised Fine-Tuning.}
The alignment process initiates with supervised fine-tuning on a 7M-sample mixture, incorporating a 5:1 ratio of general instructions to multilingual data. We adopt a three-phase curriculum: (i) basic instruction following, (ii) translation for Southeast Asian (ASEAN) languages, and (iii) integrated translation and reasoning tasks. This progression allows the model to incrementally master complex multilingual interactions. Preference rewards are derived from a reward model trained across ten categories, including accuracy, fluency, and code-mixing adherence (Appendix~\ref{sec:appendix-rm-errors}).

\paragraph{Variable Entropy Alignment.}
The VEPO integrates a clipped surrogate loss with a dynamic entropy regularization mechanism:
{\scriptsize
\begin{align}
\label{eq:vepo_objective}
\mathcal{J}_{\text{VEPO}}(\theta,\tau)
&= \mathbb{E}_{(q,a)\sim\mathcal{D},\;\{o_i\}_{i=1}^G\sim\textcolor{red}{\pi_{\theta_{\text{old}}}^\tau}}\Bigg[ 
    \frac{1}{\sum \textcolor{red}{|o_i|}}\sum_{i=1}^{G}\sum_{t=1}^{|o_i|} \nonumber\\
&\quad \min\Big( \textcolor{red}{r_{i,t}^\tau(\theta)}\textcolor{red}{\hat{A}_{i,t}^{(\text{micro}+H)}}, \nonumber\\
&\quad \operatorname{clip}\!\big(\textcolor{red}{r_{i,t}^\tau(\theta)},1-\epsilon_{\text{low}},1+\epsilon_{\text{high}}\big)\textcolor{red}{\hat{A}_{i,t}^{(\text{micro}+H)}}\Big)
\Bigg]  \nonumber \\
&\quad + \textcolor{red}{\beta \,\mathbb{E}_{(q,a),\,\{o_i\}\sim\pi_{\theta_{\text{old}}}^\tau}\Bigg[
    \frac{1}{\sum |o_i|}\sum_{i,t}
    H\!\big(\pi_\theta(\cdot\mid q,o_{i,<t})\big)
\Bigg]}
\end{align}}
where $r_{i,t}^{\tau}(\theta)$ and $\hat{A}_{i,t}^{(\text{micro}+H)}$ are defined as:
{\tiny
\begin{alignat*}{2}
r_{i,t}^{\textcolor{red}{\tau}}(\theta) &= \frac{\pi_\theta^{\textcolor{red}{\tau}}(o_{i,t}\mid q,o_{i,<t})}{\pi_{\theta_{\text{old}}}^{\textcolor{red}{\tau}}(o_{i,t}\mid q,o_{i,<t})}, \\
\hat{A}_{i,t}^{(\text{micro}+H)} &= \sg\left( \frac{R_{i,t} - \text{mean}_{\text{group}}}{\textcolor{red}{\text{std}_{\text{microbatch}}} + \epsilon} \right) \cdot \textcolor{red}{\big(1 + \alpha \sg(H) \gamma^t\big)}.
\end{alignat*}}
The VEPO alignment framework incorporates five pivotal innovations to stabilize and refine multilingual policy optimization:

\textbf{(1) Temperature-Consistent Ratios.} Unlike standard RLHF implementations that introduce distribution mismatch between sampling ($\pi^\tau$) and loss computation ($\pi^{1.0}$), VEPO strictly aligns the temperature $\tau$ across both stages. This preserves the \textit{martingale property} essential for critic-free architectures like GRPO; we empirically observed that neglecting this consistency leads to biased gradient directions and premature training collapse in low-resource regimes. 
\textbf{(2) Token-Level Contribution Balancing.} Inspired by DAPO \citep{yu2025dapoopensourcellmreinforcement}, we employ token-level normalization to ensure that samples of varying lengths contribute proportionally to the gradient. This mechanism effectively suppresses the verbosity bias inherent in sequence-level objectives, which is critical for maintaining concise and accurate translations.
\textbf{(3) Position-Aware Entropy Scheduling.} We introduce a dynamic entropy multiplier $(1 + \alpha H \gamma^t)$ that decays with output length. This ensures a high-entropy exploration space in initial decoding stages to capture diverse linguistic mappings, while gradually enforcing lower entropy to prioritize literal fidelity and structural precision as the sequence progresses.
\textbf{(4) Constraint-Driven Trajectory Filtering.} Integrated with the RLVR module, we implement a rigorous filtering strategy to prune rollout samples that exhibit linguistic pathologies, such as intra-sentence code mixing, sequence length redundancy, or "off-target" language drift. This selective optimization ensures that the policy aligns exclusively with high-quality, constraint-compliant trajectories, thereby enhancing output robustness in sensitive low-resource domains.
\textbf{(5) Communication-Efficient Advantage Normalization.} Following the REINFORCE, we normalize advantages using the group-wise mean but compute the standard deviation at the \textit{micro-batch} level. This design choice significantly reduces inter-node synchronization and collective communication bandwidth overhead, enabling high-throughput distributed training while maintaining sufficient gradient stability for actor-only updates.

\paragraph{Optimization Recipe.}
Details the VEPO iteration process. The policy collects trajectories using temperature-controlled sampling, applies RLVR-based filtering to ensure structural compliance, and updates parameters via an entropy-aware advantage estimator.
\FloatBarrier
\begin{algorithm}[!h]
    \small
    \label{alg:vepo}
    \begin{algorithmic}[1]
    \REQUIRE Initial policy $\theta_0$, temperature $\tau$, entropy $\{\beta, \gamma, \alpha\}$, clipping $\epsilon=0.2$
    \STATE Initialize $\theta \gets \theta_0$
    \WHILE{not converged}
        \STATE Sample a micro-batch of prompts $\{q_j\}_{j=1}^M \sim \mathcal{D}$
        \FOR{each prompt $q_j$}
            \STATE \textcolor{red}{\% RLVR Filtering:}
            \STATE Sample $K$ candidates $\{o_k\}_{k=1}^K \sim \pi_{\theta_{\text{old}}}^\tau(\cdot|q_j)$
            \STATE Filter candidates using $R_{\text{RLVR}}$
            \STATE Select top-$G$ compliant hypotheses $\{o_i\}_{i=1}^G$
            \STATE Compute token rewards $R_{i,t}$ and group mean $\mu_{j} = \text{mean}(\{R_{i,t}\})$
        \ENDFOR
        \STATE \textcolor{red}{\% Comm.-Efficient Normalization:}
        \STATE $\sigma_{\text{m-bsz}} \gets \text{std}(\{R_{i,t} \text{ for all } j, i \text{ in local micro-batch}\})$
        \FOR{each token $o_{i,t}$ in micro-batch}
            \STATE \textcolor{red}{\% Position-Aware Entropy \& Temperature Consistency:}
            \STATE $\hat{A}_{i,t} \gets \sg \big( \frac{R_{i,t} - \mu_{j}}{\sigma_{\text{m-bsz}} + \epsilon} \big) (1 + \alpha \sg(H) \gamma^t)$
            \STATE $r_{i,t}^\tau \gets \pi_\theta^\tau(o_{i,t}) / \pi_{\theta_{\text{old}}}^\tau(o_{i,t})$
            \STATE $L_{i,t}^{\text{clip}} \gets \min\!\big(r_{i,t}^\tau \hat{A}_{i,t},\, \operatorname{clip}(r_{i,t}^\tau, 1-\epsilon, 1+\epsilon)\hat{A}_{i,t}\big)$
        \ENDFOR
        \STATE $N \gets \text{total tokens in micro-batch}$
        \STATE $\mathcal{L}_{\text{VEPO}} \gets -\frac{1}{N}\sum L_{i,t}^{\text{clip}} - \beta \frac{1}{N}\sum H(\pi_\theta)$
        \STATE $\theta \gets \theta - \eta \nabla_\theta \mathcal{L}_{\text{VEPO}}$
    \ENDWHILE
    \RETURN $\theta$
    \end{algorithmic}
\end{algorithm}
The core philosophy of the VEPO is to sustain a high entropy exploration space within a controlled range throughout the reinforcement learning process. Unlike conventional alignment objectives that often lead to premature mode collapse resulting in monotonous or overly literal outputs. Our approach encourages the model to maintain a rich probability distribution across potential candidate tokens. 

This high entropy state is particularly crucial for models that already possess strong multilingual foundations; it provides the necessary stochasticity to move beyond rigid structural mappings and explore more nuanced, paraphrastic expressions. Under the refined guidance of the Reward Model, the policy is optimized to maximize rewards not only for semantic accuracy but also for stylistic elegance. Consequently, VEPO facilitates the emergence of "liberal translation" capabilities, allowing the model to generate translations that are both contextually accurate and linguistically sophisticated, effectively capturing the underlying intent rather than merely performing surface-level word substitution.

\section{Experiments}
\label{sec:experiments}
In this section, we present a comprehensive empirical evaluation of the Variable Entropy Policy Optimization framework across three critical dimensions. We conduct an \textbf{Algorithmic Comparison and Mechanism Analysis} (Sec.~\ref{sec:ablation}), dissecting the entropy dynamics and policy stability of VEPO relative to standard reinforcement learning baselines. We evaluate \textbf{Length Control Strategies} (Sec.~\ref{sec:overlong-analysis}), contrasting VEPO's intrinsic regulation with extrinsic penalty mechanisms to address verbosity bias. We perform \textbf{Translation Performance Benchmarking and Generalization Analysis}, validating translation quality across 90 FLORES-200 directions on BLEU, COMET, and chrF metrics while ensuring the preservation of general reasoning capabilities.

\FloatBarrier

\subsection{Constraint Verification Performance}
\label{sec:rlvr-evaluation}

To quantify the reliability of the RLVR framework \citep{wen2025reinforcementlearningverifiablerewards}, we systematically evaluate the inference-time performance of the VEPO backbone along four critical constraint dimensions: language consistency, sequence length, structural formatting, and code-mixing adherence. It is important to highlight that while the foundational Qwen2.5-7B-Instruct model exhibits a relatively low performance floor in satisfying these rigid deterministic constraints, our evaluation results summarized in Table~\ref{tab:rlvr_constraints} demonstrate that VEPO successfully bridges this gap, achieving high adherence rates even in the most challenging structural scenarios.
\FloatBarrier
\begin{table}[!ht]
    \centering
    \caption{\small Constraint verification performance (\%). Our model (top) consistently achieves superior alignment across deterministic structural constraints.}
    \label{tab:rlvr_constraints}
    \renewcommand{\arraystretch}{1.2}
    \setlength{\tabcolsep}{6pt}
    \scriptsize
    \begin{threeparttable}
    \resizebox{\columnwidth}{!}{%
    \begin{tabular}{@{}lccccc@{}}
        \toprule
        \rowcolor{gray!15}
        \textbf{Model} & \textbf{Lang.} & \textbf{Length} & \textbf{Format} & \textbf{Mixing} & \textbf{Overall} \\
        \midrule
        \rowcolor{blue!5}
        \textbf{VEPO-7B (Ours)} & \textbf{97.8} & 99.2 & \textbf{92.15} & 92.3 & \textbf{95.3} \\
        Qwen2.5-7B-Instruct & 92.0 & 97.0 & 51.8 & 62.8 & 75.9 \\
        Gemma3-27B-IT & 97.4 & 91.6 & 42.1 & 90.9 & 80.5 \\
        Qwen-MT-Plus & 97.6 & \textbf{99.8} & 82.5 & 94.8 & 93.6 \\
        Seed-X-PPO-7B & 97.6 & 79.8 & 79.0 & 90.3 & 86.6 \\
        DeepSeek-V3 & 95.4 & 95.7 & 67.6 & 95.0 & 88.4 \\
        Hunyuan-MT-7B & 91.8 & 90.7 & 71.1 & \textbf{96.2} & 87.4 \\
        \bottomrule
    \end{tabular}%
    }
    \end{threeparttable}
\end{table}

As shown in Table~\ref{tab:rlvr_constraints}, VEPO achieves an overall constraint satisfaction rate of 95.3\%, significantly outperforming general-purpose chat models and specialized translation systems. These results validate the efficacy of the RLVR layer in enforcing deterministic structural requirements without compromising semantic fidelity.

\subsection{Algorithmic Comparison and Mechanism Analysis}
\label{sec:ablation}

To systematically dissect the interplay between advantage estimation, regularization constraints, and policy stability, we conducted a comprehensive benchmark spanning 18 experimental configurations. We evaluated VEPO against five baselines: PPO \citep{schulman2017proximalpolicyoptimizationalgorithms}, GRPO \citep{shao2024deepseekmathpushinglimitsmathematical}, DAPO \citep{yu2025dapoopensourcellmreinforcement}, RLOO\citep{ahmadian2024basicsrevisitingreinforcestyle}, and Reinforce++ \citep{hu2025reinforcestabilizingcriticfreepolicy}, across three KL divergence regimes: K2, K3 and none-KL.

\FloatBarrier
\begin{figure}[!hbp]
    \centering
    \includegraphics[width=\columnwidth]{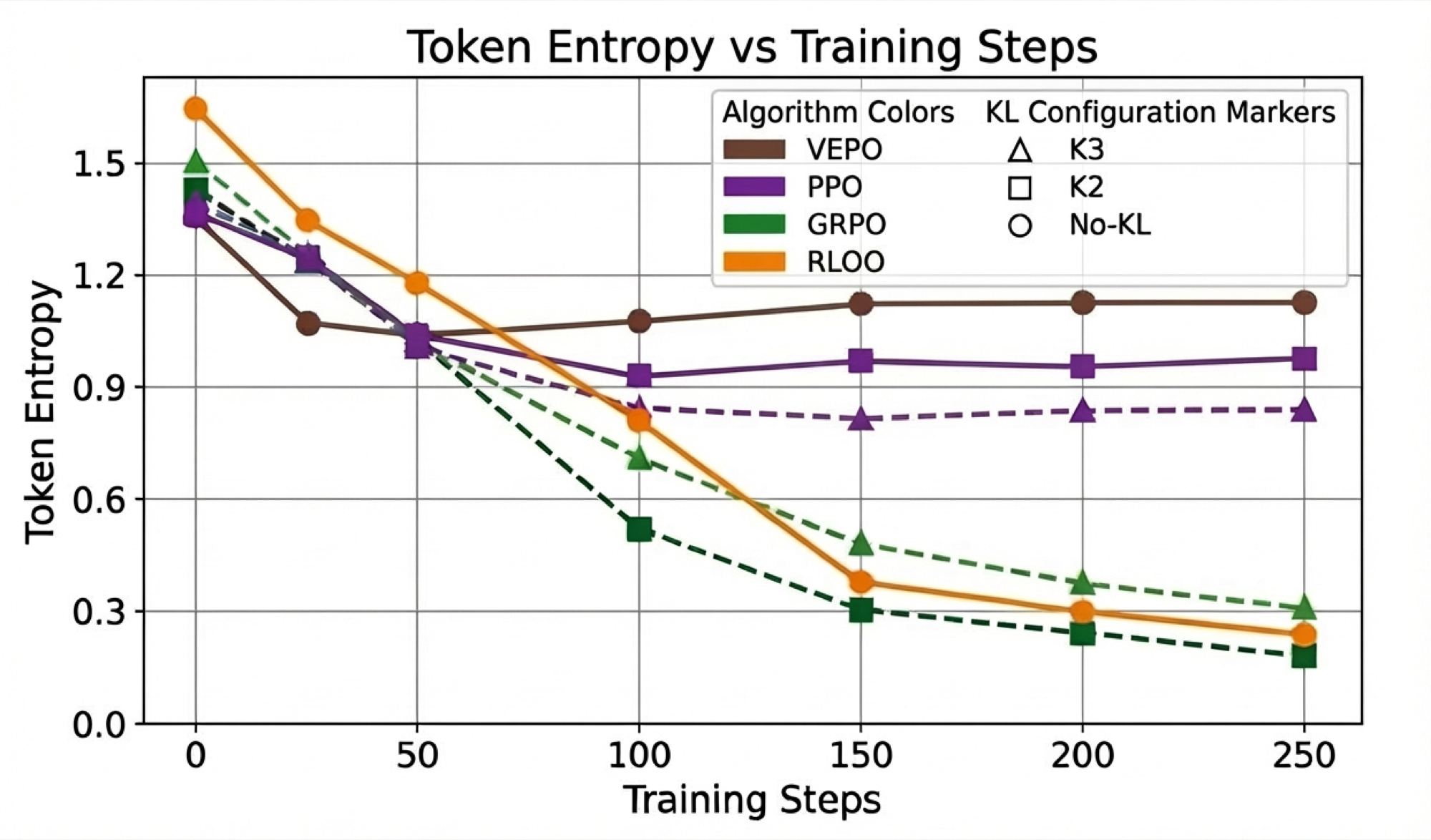}
    \captionsetup{font=small}
    \caption{Policy entropy dynamics across RL algorithms and KL regimes.}
    \label{fig:entropy_loss}
\end{figure}

Figure~\ref{fig:entropy_loss} illustrates a stark divergence in training dynamics contingent on the presence of a value baseline. We observe that GRPO undergoes rapid entropy collapse, converging prematurely to deterministic modes. This suggests that in the absence of variance-reduction mechanisms, group-relative advantages tend to over-reinforce local optima, leading to catastrophic exploitation. In contrast, PPO variants leverage the value critic to smooth the optimization landscape, maintaining relative stability.

Notably, VEPO exhibits superior entropy retention in the unconstrained No-KL regime, outperforming even critic-aided PPO. This indicates that VEPO's depth-dependent entropy modulation provides a more direct form of policy regularization than external KL constraints or value functions. By dynamically balancing exploration and exploitation through its variable entropy objective, VEPO intrinsically prevents distributional collapse, establishing a robust alternative to heavy critic-based architectures.

\subsection{Length Control Strategy Analysis}
\label{sec:overlong-analysis}

\FloatBarrier
\begin{figure}[!h]
    \centering
    \includegraphics[width=\columnwidth]{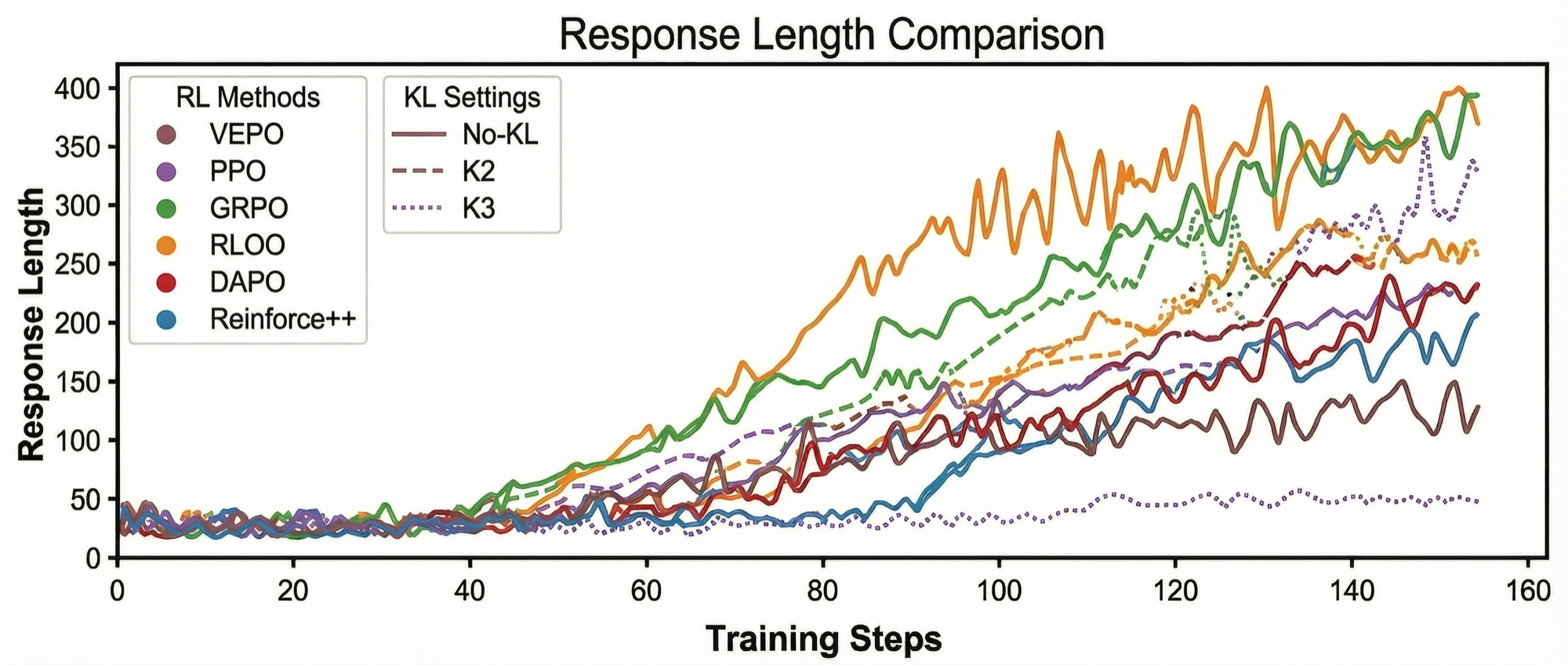}
    \captionsetup{font=small}
    \caption{Response length stability across six RL algorithms and three KL regimes (18 independent runs).}
    \label{fig:length_comparison_all}
\end{figure}

The ability to regulate response length is a crucial marker of RL stability, as models often exploit "verbosity bias" to inflate reward signals—a classic form of reward hacking. To provide a rigorous evaluation, we conducted an ablation study spanning 18 independent experimental runs, comparing six algorithms across three KL regimes under strictly identical hyperparameter settings. As illustrated in Figure~\ref{fig:length_comparison_all}, the algorithms are distinguished by color: brown for VEPO, purple for PPO, green for GRPO, orange for RLOO, red for DAPO, and blue for Reinforce++. The line styles denote the KL configurations, with solid, dashed, and dotted lines representing the No-KL, K2, and K3 regimes, respectively. The results demonstrate that while PPO remains stable due to its value baseline, VEPO is the only other framework that consistently prevents response redundancy across all KL configurations. Even in the unconstrained No-KL regime (solid lines), where baselines like RLOO and GRPO exhibit severe length drift, VEPO maintains a stable trajectory, effectively mitigating reward hacking through its intrinsic policy dynamics.

\FloatBarrier
\begin{figure}[!h]
    \centering
    \includegraphics[width=0.9\columnwidth,keepaspectratio]{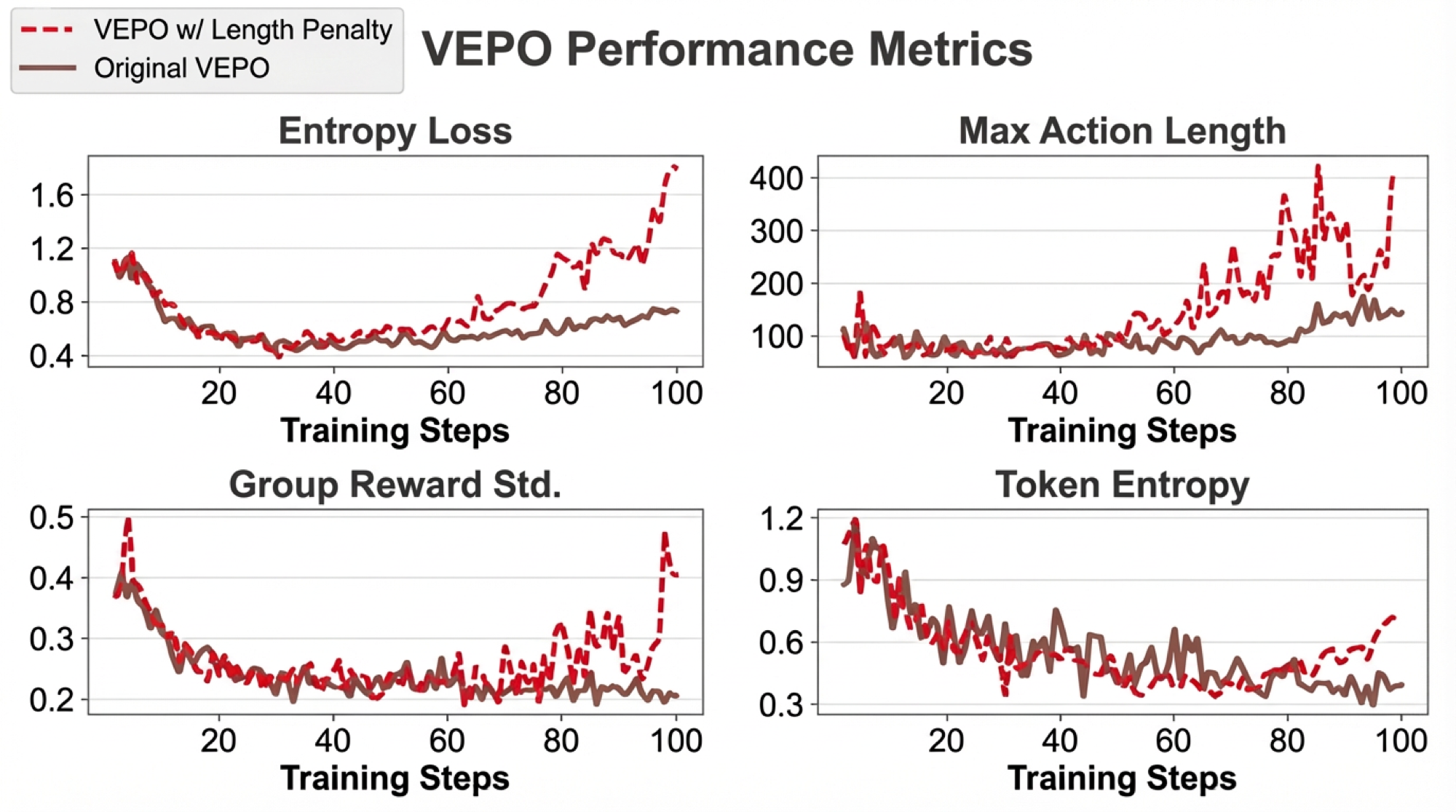}
    \captionsetup{font=small}
    \caption{Sensitivity analysis of explicit length penalties. External constraints induce reward instability and entropy oscillations, leading to training divergence.}
    \label{fig:vepo-overlong}
\end{figure}

To further dissect this intrinsic stability, Figure~\ref{fig:vepo-overlong} isolates the impact of DAPO's overlength penalties. While explicit penalties can truncate sequence length, they often provoke severe entropy oscillations and eventual training collapse. VEPO’s success in achieving length regulation without such destabilizing extrinsic constraints suggests that its variable entropy objective functions as a self-regulating mechanism. By penalizing the "certainty" of over-extended sequences via depth-dependent modulation, VEPO suppresses pathological verbosity through intrinsic policy dynamics rather than extrinsic reward manipulation.

\FloatBarrier
\subsection{Performance Translation Benchmarking and Generalization Analysis}

\begin{figure}[H]
    \centering
    \includegraphics[width=0.90\columnwidth]{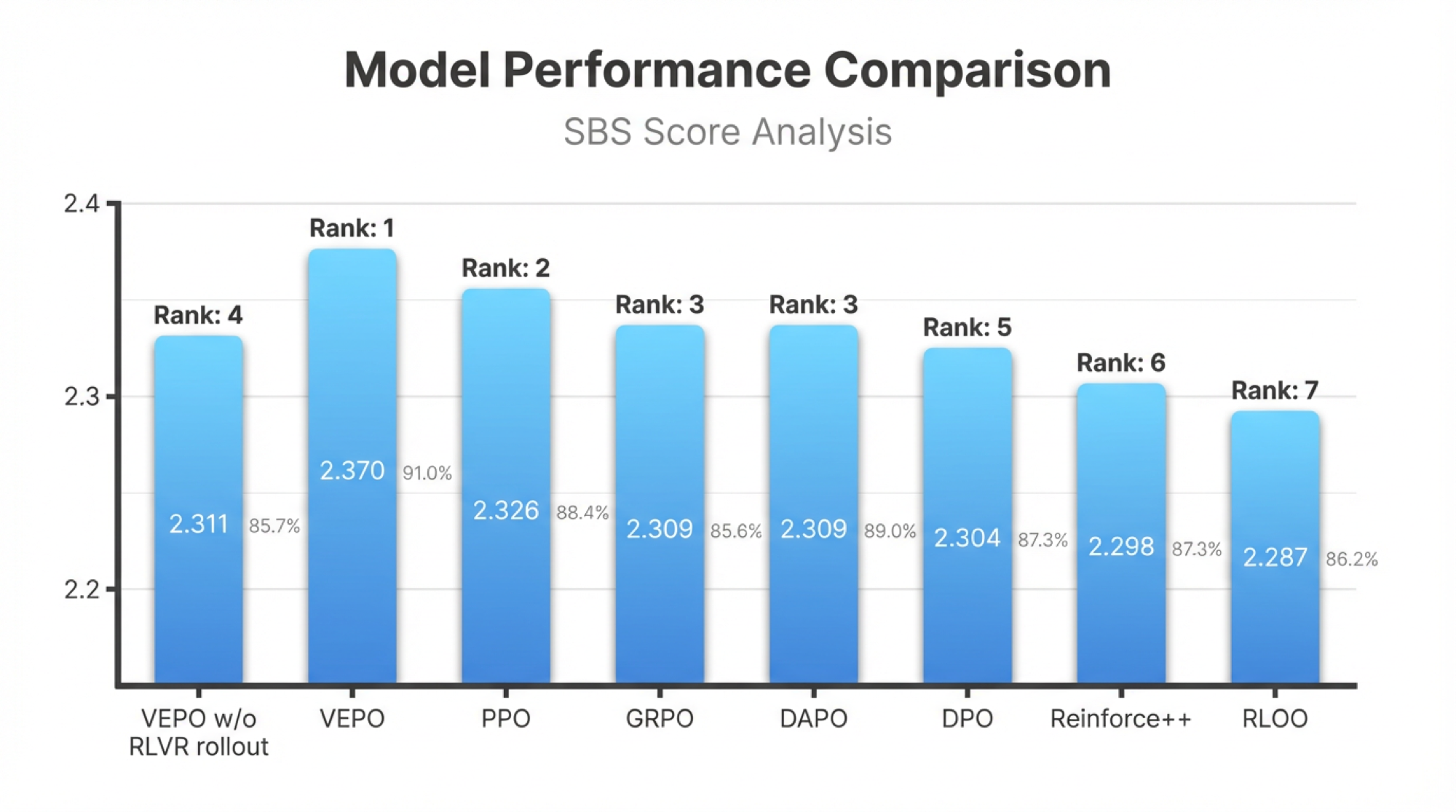}
    \caption{\small Results of pairwise ranking comparisons from professional human evaluations of translation samples.}
    \label{fig:kkk}
\end{figure}

To further validate the practical utility of our framework, we conducted a rigorous human evaluation involving professional multilingual experts. Following a side-by-side evaluation protocol similar to reward model ranking logic, experts performed blind comparisons of translation outputs across various language pairs. Annotators assigned scores on a 0--3 scale, where the reported percentage represents the proportion of samples receiving a high-quality rating (score 2 or 3). The evaluation criteria encompass both semantic accuracy and paraphrastic fidelity, ensuring that the model not only captures literal meaning but also preserves naturalness and idiomatic expression. As illustrated in Figure~\ref{fig:kkk}, the VEPO algorithm demonstrates superior performance across all compared baselines, achieving the highest human preference scores while maintaining high translation quality.

\begin{figure*}[htbp]
    \centering
    \includegraphics[width=0.90\textwidth]{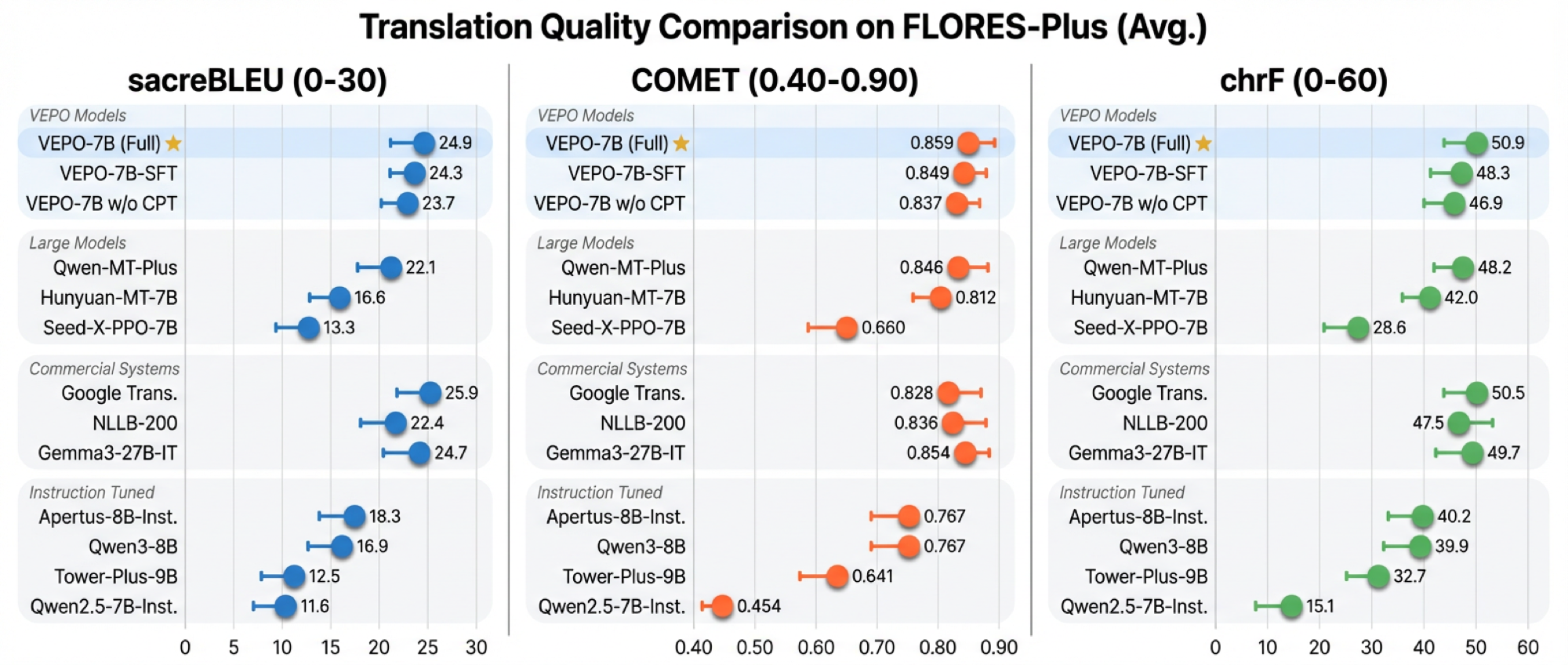}
    \caption{Overall translation performance comparison (Avg.) on FLORES-200 across sacreBLEU, COMET, and chrF metrics.}
    \label{fig:main_metrics_visual}
\end{figure*}


\begin{table}[htbp]
    \centering
    \caption{\small Performance on FLORES-200, COMET-22, chrF}
    \label{tab:translation_results}
    \renewcommand{\arraystretch}{1.2}
    \setlength{\tabcolsep}{1.2pt}
    \scriptsize
    \begin{threeparttable}
    \resizebox{\columnwidth}{!}{%
    \begin{tabular}{@{}lcccc@{}}
        \toprule
        \rowcolor{gray!15}
        \textbf{Model} & \textbf{X$\rightarrow$E} & \textbf{E$\rightarrow$X} & \textbf{X$\rightarrow$X} & \textbf{Avg.} \\
        \midrule
        \rowcolor{gray!5}
        \multicolumn{5}{@{}l}{\textit{Multilingual Chat Models}} \\
        Gemma3-27B-IT & 36.8/.882/62.9 & 30.7/.875/54.2 & 22.3/.847/47.5 & 24.7/.854/49.7 \\
        Qwen3-8B & 31.1/.862/58.5 & 23.3/.799/46.4 & 14.4/.752/36.8 & 16.9/.767/39.9 \\
        \underline{\textcolor{red}{Qwen2.5-7B-Inst.}} & \underline{\textcolor{red}{24.8/.498/19.4}} & \underline{\textcolor{red}{17.4/.455/15.6}} & \underline{\textcolor{red}{9.2/.449/14.5}} & \underline{\textcolor{red}{11.6/.454/15.1}} \\
        Apertus-8B-Inst. & 32.5/.870/60.5 & 25.7/.802/46.6 & 15.6/.750/36.9 & 18.3/.767/40.2 \\
        Tower-Plus-9B & 28.2/.825/54.7 & 18.3/.671/37.7 & 9.8/.615/29.3 & 12.5/.641/32.7 \\
        \midrule
        \rowcolor{gray!5}
        \multicolumn{5}{@{}l}{\textit{Translation-Focused Models}} \\
        Qwen-MT-Plus & 34.0/.881/60.1 & 29.6/.869/53.4 & 19.6/.839/45.8 & 22.1/.846/48.2 \\
        Seed-X-PPO-7B & 25.9/.786/51.6 & 22.6/.708/36.3 & 10.5/.638/24.7 & 13.3/.660/28.6 \\
        Hunyuan-MT-7B & 24.6/.839/55.2 & 23.4/.862/48.4 & 14.8/.802/39.5 & 16.6/.812/42.0 \\
        \midrule
        \rowcolor{gray!5}
        \multicolumn{5}{@{}l}{\textit{Translation-Focused without LLM Models}} \\
        Google Trans. & \textbf{41.2}/\textbf{.884}/\textbf{65.7} & 32.2/.842/55.7 & \textbf{23.2}/.820/47.9 & \textbf{25.9}/.828/50.5 \\
        NLLB-200 & 38.1/.875/63.0 & 28.3/.845/52.1 & 19.7/.830/44.9 & 22.4/.836/47.5 \\
        \midrule
        \rowcolor{gray!5}
        \multicolumn{5}{@{}l}{\textit{Our Models (VEPO Framework)}} \\
        VEPO-7B w/o CPT & 33.3/.862/56.8 & 31.7/.863/51.8 & 21.4/.822/43.6 & 23.7/.837/46.9 \\
        \rowcolor{gray!12}\quad $\Delta$ (VEPO $-$ w/o CPT) & +2.8/+.019/+5.9 & +1.0/+.019/+4.4 & +1.7/+.032/+5.2 & +1.2/+.022/+4.1 \\
        VEPO-7B-SFT & 35.4/.875/59.8 & 32.0/.875/52.9 & 22.7/.839/44.5 & 24.3/.849/48.3 \\
        \rowcolor{gray!12}\quad $\Delta$ (VEPO $-$ SFT) & +0.7/+.006/+2.9 & +0.7/+.007/+3.3 & +0.4/+.015/+4.3 & +0.6/+.010/+2.7 \\
        \rowcolor{blue!5}
        \textbf{VEPO-7B (Full)} & 36.1/.881/62.7 & \textbf{32.7}/\textbf{.882}/\textbf{56.2} & 23.1/\textbf{.854}/\textbf{48.8} & 24.9/\textbf{.859}/\textbf{50.9} \\
        \bottomrule
    \end{tabular}%
    }
    \end{threeparttable}
\end{table}

We evaluate VEPO on the FLORES-200 dataset (covering 90 ASEAN language directions) using sacreBLEU \citep{post-2018-call}, COMET \citep{rei-etal-2022-comet}, and chrF \citep{popovic-2015-chrf}, while concurrently assessing general reasoning capabilities via BBH \citep{suzgun2022challengingbigbenchtaskschainofthought}, CMMLU \citep{li2024cmmlumeasuringmassivemultitask}, HellaSwag \citep{zellers-etal-2019-hellaswag}, and MMLU \citep{hendrycks2021measuringmassivemultitasklanguage}. All evaluations are conducted within the 7B-parameter regime to ensure a rigorous comparison against leading multilingual and translation-specific baselines.

As summarized in Table~\ref{tab:translation_results} and Figure~\ref{fig:main_metrics_visual}, VEPO-7B (Full) establishes a new state-of-the-art for open-source 7B models across all translation directions. Notably, it achieves an Average BLEU of 24.9, effectively narrowing the performance gap with commercial systems like Google Translate and significantly outperforming the much larger Gemma3-27B-IT in E$\rightarrow$X and X$\rightarrow$X scenarios. The ablation analysis ($\Delta$) underscores the synergistic effect of our framework: tokenizer-driven continued pre-training (CPT) yields substantial gains in linguistic coverage, while the VEPO policy optimization further refines semantic fidelity beyond standard SFT baselines.

\begin{table}[htbp]
    \centering
    \caption{\small General reasoning benchmarks (\%). \underline{\textcolor{red}{Red}}: baseline.}
    \label{tab:general_capabilities}
    \renewcommand{\arraystretch}{1.2}
    \setlength{\tabcolsep}{5pt}
    \scriptsize
    \begin{tabular}{@{}lcccc@{}}
        \toprule
        \rowcolor{gray!15}
        \textbf{Model} & \textbf{BBH} & \textbf{CMMLU} & \textbf{HellaS.} & \textbf{MMLU} \\
        \midrule
        \rowcolor{gray!5}
        \multicolumn{5}{@{}l}{\textit{Multilingual Chat Models}} \\
        Gemma3-27B-IT & 55.9 & 55.9 & 55.9 & \textbf{56.0} \\
        Qwen3-8B & \textbf{63.8} & 60.8 & 26.0 & 51.3 \\
        \underline{\textcolor{red}{Qwen2.5-7B-Inst.}} & \underline{\textcolor{red}{54.4}} & \underline{\textcolor{red}{\textbf{64.1}}} & \underline{\textcolor{red}{\textbf{85.2}}} & \underline{\textcolor{red}{40.9}} \\
        Apertus-8B-Inst. & 49.2 & 45.3 & 64.2 & 45.2 \\
        Tower-Plus-9B & 40.4 & 57.2 & 73.1 & 42.1 \\
        \midrule
        \rowcolor{gray!5}
        \multicolumn{5}{@{}l}{\textit{Our Models}} \\
        VEPO-7B w/o CPT & 59.7 & 62.0 & 83.5 & 47.6 \\
        \rowcolor{gray!12}\quad $\Delta$ (VEPO $-$ w/o CPT) & +1.2 & +1.2 & +1.7 & +0.9 \\
        VEPO-7B-SFT & 59.6 & 61.4 & 83.7 & 47.2 \\
        \rowcolor{gray!12}\quad $\Delta$ (VEPO $-$ SFT) & +1.3 & +1.8 & +1.5 & +1.3 \\
        \rowcolor{blue!5}
        \textbf{VEPO-7B (Full)} & 60.9 & 63.2 & \textbf{85.2} & 48.5 \\
        \bottomrule
    \end{tabular}
\end{table}

Furthermore, Table~\ref{tab:general_capabilities} demonstrates that our translation-centric optimization does not incur the "catastrophic forgetting" typically associated with model specialization. VEPO-7B maintains or even enhances performance on general reasoning benchmarks, suggesting that our variable entropy mechanism successfully preserves the underlying instruction-following capacity while mastering low-resource scripts. This dual proficiency positions VEPO as a robust foundation for multilingual applications where both literal accuracy and general intelligence are required. Detailed per-language breakdowns are provided in Appendices~\ref{sec:appendix-blue}--\ref{sec:appendix-chrf}.

Training configurations and evaluation metrics are detailed in Appendix~\ref{sec:appendix-exp-setup}.

\section{Analysis}
\label{sec:analysis}

We provide a rigorous theoretical derivation of how the VEPO framework structurally transforms the policy optimization landscape to enable paraphrastic exploration.

\subsection{Why High Entropy Facilitates Paraphrastic Translation}

Machine translation is intrinsically a \emph{one-to-many} mapping: for a given source sentence $q$,
there exists a non-trivial equivalence class of acceptable translations
\[
\Omega_{q,\delta} \triangleq \{ o \;:\; R(q,o) \ge R^*(q) - \delta \},
\]
where $R(q,o)$ denotes a semantic fidelity reward and $\delta$ controls near optimality.
In idiomatic and colloquial domains, $R(q,o)$ is typically \emph{approximately flat} over $\Omega_{q,\delta}$,
creating a broad paraphrastic plateau rather than a single sharp optimum.

Consider the maximum entropy objective over sequence distributions:
\[
\max_{\pi(\cdot\mid q)} \;\; 
\mathbb{E}_{o\sim\pi(\cdot\mid q)}[R(q,o)] + \beta H(\pi(\cdot\mid q)).
\]
The stationary solution is the Gibbs distribution
\[
\pi^*(o\mid q)\propto \exp\!\left(\frac{R(q,o)}{\beta}\right).
\]
Consequently, when $R(q,o)$ is nearly constant on $\Omega_{q,\delta}$ and significantly lower outside,
$\pi^*(\cdot\mid q)$ approaches a near uniform distribution over $\Omega_{q,\delta}$,
allocating non-zero probability mass to multiple paraphrastic realizations.
Therefore, entropy regularization is not merely ``uncertainty'': it increases the \emph{support coverage}
over the paraphrastic equivalence class, enabling the policy to represent and discover diverse yet faithful translations.

\subsection{Entropy-Regularized Optimization Geometry}
In token level policies, mode collapse corresponds to the policy approaching the boundary of the probability simplex, where exploration vanishes and optimization becomes ill conditioned.
For a categorical policy with probabilities $p$, the Fisher information takes the form
$\mathcal{G}=\mathrm{diag}(p)-pp^\top$, whose eigenvalues shrink toward $0$ as $p$ becomes deterministic.
The entropy bonus $\beta H(\pi_\theta)$ counteracts this boundary attraction by penalizing degenerate distributions,
maintaining a non-trivial probability floor for paraphrastic alternatives and improving numerical conditioning.
\textbf{Position-dependent exploration.}
Paraphrastic divergence in translation predominantly happens early in decoding,
when the model selects a lexicalized phrase or idiomatic template.
We thus apply a position-decayed coefficient $\gamma^t$ to concentrate exploration on early tokens,
while allowing later tokens to collapse toward a coherent realization of the chosen paraphrastic template.
This yields a controlled exploration--commitment dynamic that increases paraphrastic coverage without inducing late-stage drift.

\subsection{Temperature Consistent Ratios for Unbiased Optimization}
VEPO samples trajectories from the tempered behavior policy $\pi_{\theta_{\text{old}}}^\tau$.
To obtain an unbiased estimator of expectations under the current tempered policy $\pi_\theta^\tau$,
the importance ratio must be computed under the \emph{same} temperature:
\[
r_t^\tau(\theta)=\frac{\pi_\theta^\tau(o_t\mid q,o_{<t})}{\pi_{\theta_{\text{old}}}^\tau(o_t\mid q,o_{<t})},
\quad
\mathbb{E}_{\pi_{\theta_{\text{old}}}^\tau}\!\left[r^\tau f\right]
=
\mathbb{E}_{\pi_\theta^\tau}[f].
\]
Decoupling the sampling temperature from the ratio temperature implicitly changes the optimized objective
and introduces estimation bias, especially in high temperature exploration regimes.

\subsection{Restoring the Paraphrastic Manifold}

To empirically validate our theoretical findings on "curvature flattening," we perform a microscopic logit analysis at critical decision points. We use the idiom "It's a piece of cake" (English $\rightarrow$ Vietnamese) as a representative case, where a model must choose between a literal translation ("mi\'{\^{e}}ng" / piece) and a paraphrastic refinement ("d\~{\^{e}}" / easy).
\begin{figure}[h]
    \centering
    \includegraphics[width=0.99\columnwidth]{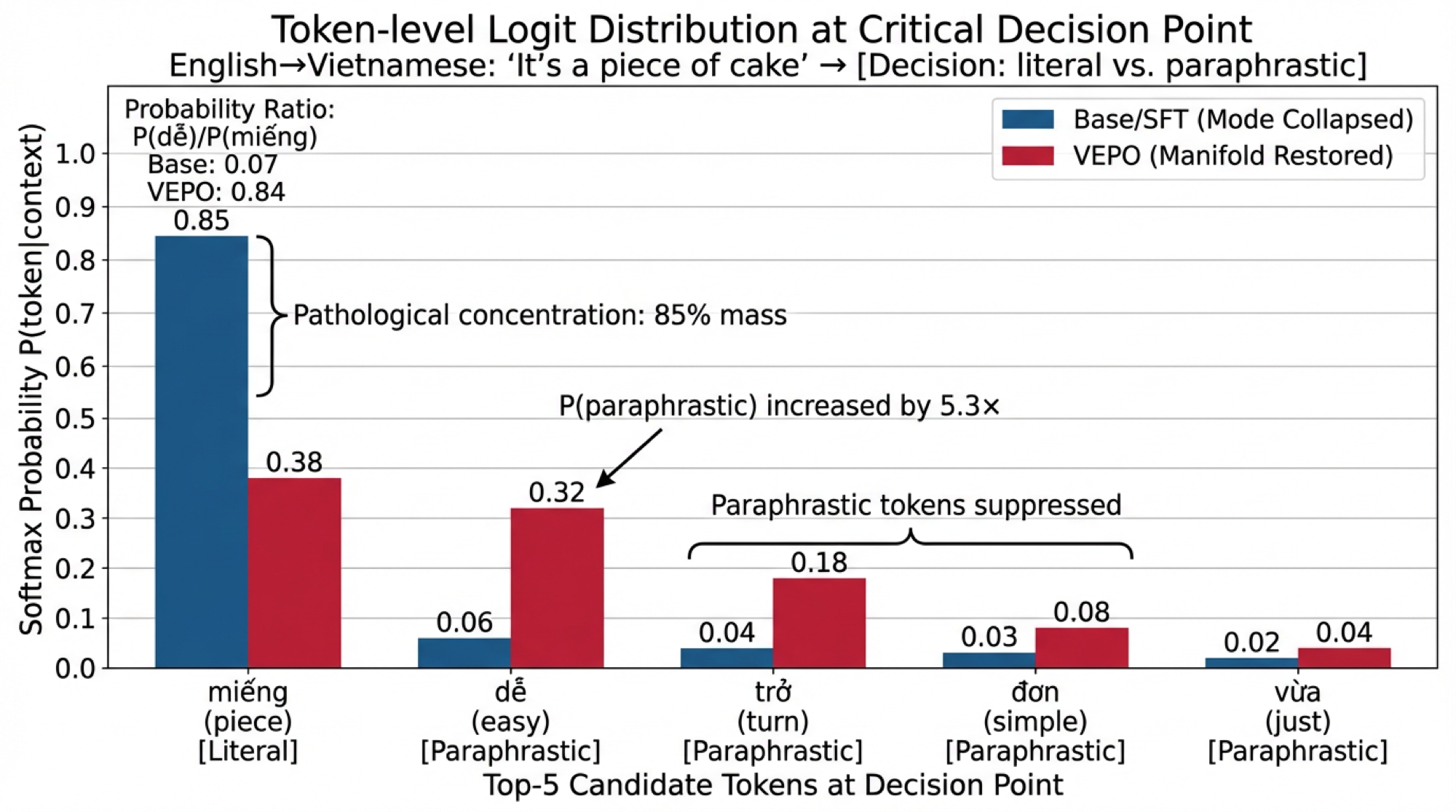}
    \captionsetup{font=small}
    \caption{\small Token level logit distribution comparison at a critical decision point. VEPO reactivates the paraphrastic manifold, allowing the model to explore refined linguistic mappings that are suppressed by mode collapse in the base SFT model.}
    \label{fig:logit_comparison}
\end{figure}
As illustrated in Figure~\ref{fig:logit_comparison}, the Base/SFT model exhibits a \textbf{pathological concentration} of probability mass ($85\%$) on the literal token "mi\'{\^{e}}ng," effectively suppressing all paraphrastic alternatives to $<6\%$. This confirms the "mode collapse" predicted in Section 6.1, where the model's Fisher metric becomes singular around the most frequent SFT patterns. 

In contrast, VEPO \textbf{reactivates the paraphrastic manifold}. By maintaining a probabilistic buffer, the mass for the refined token "d\~{\^{e}}" (easy) increases from $6\%$ to $32\%$ (a $5.3\times$ boost), and the probability ratio $P(\text{d\~{\^{e}}})/P(\text{mi\'{\^{e}}ng})$ shifts from a negligible $0.07$ to a competitive $0.84$. This redistribution does not represent "uncertainty," but rather \textbf{expressive bandwidth}: the model now recognizes multiple valid linguistic paths. This restored manifold allows the Reward Model to successfully guide the policy toward more "liberal" and "refined" translations. Consequently, VEPO trained models excel at linguistic refinement—preferring contextually elegant expressions over rigid word-for-word substitutions—without sacrificing semantic fidelity.
\section{Conclusion}

In this paper, we introduced \textbf{Variable Entropy Policy Optimization}, a comprehensive framework for adapting foundation models to low-resource linguistic environments. Our approach systematically addresses the primary bottlenecks in multilingual modeling through targeted tokenizer expansion, balanced continued pre-training, and entropy-aware reinforcement learning. By integrating Reinforcement Learning with Verifiable Rewards, we enforce deterministic structural constraints directly within the optimization loop, effectively mitigating common failure modes such as sequence inflation and markup corruption. Empirical results across 90 FLORES-200, COMET-22 and chrF directions demonstrate that VEPO achieves state-of-the-art translation performance while preserving robust general-purpose capabilities. 

Future research will focus on refining reward models for high-fidelity paraphrastic evaluation and exploring advanced RL methodologies to better handle linguistic diversity. The principles of dynamic entropy modulation and verifiable alignment presented here offer a promising foundation for building more robust, inclusive, and expressive multilingual models.
\section*{Impact Statement}

This paper presents Variable Entropy Policy Optimization (VEPO), a framework dedicated to improving the performance and reliability of Large Language Models in low-resource linguistic contexts. By systematically addressing bottlenecks such as tokenization inefficiency and reinforcement learning instabilities, our work contributes to the development of more inclusive and accessible AI technologies. 

The primary impact of this research is the potential to bridge the digital divide for speakers of underrepresented languages, enabling them to benefit more effectively from state-of-the-art machine translation and instruction-following capabilities. Furthermore, by integrating Reinforcement Learning with Verifiable Rewards (RLVR) to enforce structural and linguistic constraints, we provide a methodology for building models that are not only more capable but also more robust and predictable in their behavior. 

Advances in translation technology can be weaponized—turbocharging cross-lingual misinformation—but the upside is far greater: deeper cross-cultural understanding, stronger support for linguistic diversity, and powerful new tools for marginalized communities. Our approach drives responsible multilingual model development by demanding both top-tier performance and precise, deterministic control over every output.

\clearpage
\bibliographystyle{icml2026}

\newpage
\appendix
\onecolumn
\section{Supplementary Material}

\subsection{Tokenizer Expansion Details}
\label{sec:appendix-vocab-expansion}

This section provides detailed tokenizer diagnostics for the expanded low resource languages, showing compression ratios and efficiency improvements achieved through vocabulary expansion.

\subsection{Detailed RLVR Formulations}
\label{sec:appendix-rlvr-formulas}

The Reward Learning with Verifiable Rewards (RLVR) framework combines semantic translation quality signals with deterministic, verifiable constraints. The complete reward function is defined as:

\begin{align}
R_{\text{RLVR}}(x,y) &= R_{\text{mt}}(x,y) + \lambda_{\text{len}}\, r_{\text{len}}(x,y) + \lambda_{\text{fmt}}\, r_{\text{fmt}}(x,y) \nonumber \\
&\quad + \lambda_{\text{lid}}\, r_{\text{lid}}(y) + \lambda_{\text{mix}}\, r_{\text{mix}}(y)
\label{eq:rlvr-full}
\end{align}

where $R_{\text{mt}}$ represents the semantic translation reward and each verifiable term is defined as follows:

\subsubsection{Length Ratio Reward}

The length ratio reward $r_{\text{len}}(x,y)$ constrains output length relative to input:

\begin{equation}
r_{\text{len}}(x,y) = 
\begin{cases}
+1, & \text{if } \rho \in [\alpha, \beta] \\
-\sigma_{\text{len}} \cdot \max(0, \rho - \beta), & \text{if } \rho > \beta \\
-\sigma_{\text{len}} \cdot \max(0, \alpha - \rho), & \text{if } \rho < \alpha
\end{cases}
\label{eq:length-ratio}
\end{equation}

where $\rho = \frac{|y|}{|x|}$ is the length ratio, $[\alpha, \beta]$ defines the acceptable ratio range (typically $[0.5, 2.0]$ for most language pairs), and $\sigma_{\text{len}} > 0$ controls the penalty strength for violations.

\subsubsection{Format Validity Reward}

For inputs containing structural tokens (HTML, Markdown, etc.), the format validity reward $r_{\text{fmt}}(x,y)$ is computed as:

\begin{equation}
r_{\text{fmt}}(x,y) = w_{\text{preserve}} \cdot f_{\text{preserve}}(x,y) - w_{\text{broken}} \cdot f_{\text{broken}}(y)
\label{eq:format-validity}
\end{equation}

where:
\begin{itemize}
\item $f_{\text{preserve}}(x,y) = \frac{|\text{struct\_tokens}(x) \cap \text{struct\_tokens}(y)|}{|\text{struct\_tokens}(x)|}$ measures the preservation ratio of structural tokens
\item $f_{\text{broken}}(y)$ counts syntax violations (unclosed tags, broken lists, malformed code fences) detected by lightweight parsers
\item $w_{\text{preserve}}, w_{\text{broken}} > 0$ are weighting factors
\end{itemize}

\subsubsection{Target Language Verification}

The language identification reward $r_{\text{lid}}(y)$ verifies that the output matches the target language $\ell_t$:

\begin{equation}
r_{\text{lid}}(y) = 
\begin{cases}
+1, & \text{if } \text{LID}(y) = \ell_t \text{ and } \text{confidence} > \theta_{\text{lid}} \\
-\eta_{\text{lid}}, & \text{otherwise}
\end{cases}
\label{eq:language-id}
\end{equation}

where $\text{LID}(\cdot)$ is a language identification model, $\theta_{\text{lid}}$ is the confidence threshold (typically 0.8), and $\eta_{\text{lid}} > 0$ is the penalty for off-target language.

\subsubsection{Code-mixing Detection}

The code-mixing reward $r_{\text{mix}}(y)$ penalizes excessive mixing of languages within the output:

\begin{equation}
r_{\text{mix}}(y) = 
\begin{cases}
0, & \text{if } p_{\text{mix}}(y) \leq \tau_{\text{mix}} \\
-\zeta_{\text{mix}} \cdot (p_{\text{mix}}(y) - \tau_{\text{mix}}), & \text{otherwise}
\end{cases}
\label{eq:code-mixing}
\end{equation}

where $p_{\text{mix}}(y)$ estimates the proportion of non-$\ell_t$ segments or scripts in $y$, $\tau_{\text{mix}}$ is the mixing tolerance threshold (typically 0.1-0.2), and $\zeta_{\text{mix}} > 0$ controls the penalty severity.

\subsubsection{Implementation Details}

All verifiable terms are designed to be computationally efficient and deterministic:

\begin{itemize}
\item \textbf{Clipping}: Each reward term is clipped to $[-C_{\max}, C_{\max}]$ to prevent gradient explosions, where $C_{\max} = 5.0$ in our implementation.
\item \textbf{Normalization}: The final reward is normalized across the batch to maintain stable training dynamics.
\item \textbf{Hyperparameters}: We use $\lambda_{\text{len}} = 0.3$, $\lambda_{\text{fmt}} = 0.2$, $\lambda_{\text{lid}} = 0.4$, $\lambda_{\text{mix}} = 0.3$ as default weights, tuned on validation sets.
\end{itemize}

The RLVR framework thus provides structured, interpretable constraints that complement learned semantic rewards while maintaining computational efficiency during training.
\subsection{Reward Model Error Categories}
\label{sec:appendix-rm-errors}

In our reward model design, we systematically categorized translation errors into ten distinct types to ensure comprehensive coverage of potential failure modes. This categorization framework was essential for training a robust reward model that could effectively identify and appropriately score various types of translation errors across multiple languages.

\begin{table}[htp]
    \centering
    \caption{Error categories for RM robustness}
    \label{tab:error_categories}
    \renewcommand{\arraystretch}{1.0}
    \scriptsize
    \begin{threeparttable}
    \begin{tabular}{l l p{6cm}}
    \toprule
    \textbf{Category}\tnote{1} & \textbf{Type} & \textbf{Description} \\
    \midrule
    \multirow{6}{*}{Quality} & Accuracy & Mistranslation, omission, over-translation, semantic misunderstanding \\
    & Off-target & Untranslated segments, wrong target language \\
    & Fluency & Natural expression, comprehensibility, professionalism \\
    & Terminology & Consistency, accuracy, omission of terms \\
    & Code-mixing & Unreasonable language mixing within sentences \\
    & Repetition & End-of-sentence, segment, instruction repetition \\
    \midrule
    \multirow{2}{*}{Instruction} & Intent & Machine misunderstanding of instructions \\
    & Leakage & Output containing or translating instructions \\
    \midrule
    Content & Explanation & Excessive explanations beyond translation \\
    \midrule
    Format\tnote{2} & Preservation & Maintaining original formatting and punctuation \\
    \bottomrule
    \end{tabular}
    \begin{tablenotes}[flushleft]
      \footnotesize
      \item[1] Categories are non-exclusive; a sample may trigger multiple error types.
      \item[2] Preservation includes punctuation, special markers, and structural tokens.
    \end{tablenotes}
    \end{threeparttable}
\end{table}

This comprehensive error categorization system enabled our reward model to provide nuanced feedback on translation quality, distinguishing between different types of errors and their relative severity. The framework was particularly important for low-resource languages where translation errors can manifest in diverse and subtle ways.
\subsection{Detailed VEPO Mathematical Derivations}
\label{sec:appendix-vepo-details}

This section provides comprehensive mathematical details for the VEPO formulation presented in the main text.

\subsubsection{Advantage Estimation and Normalization}

The VEPO advantage estimator is constructed as:

\begin{equation}
\hat{A}_{i,t} = \sg\left( \frac{R_{i,t} - B_t(q)}{\text{std}_{\text{microbatch}}(\{R_{j,t}\}_{j=1}^G) + \epsilon} \right) \cdot (1 + \alpha \cdot \sg(H) \cdot \gamma^t)
\label{eq:vepo-advantage-detailed}
\end{equation}

where $B_t(q) = \text{mean}_{\text{group}}(\{R_{j,t}\}_{j=1}^G)$ is the groupwise baseline and $\sg(\cdot)$ denotes the stop-gradient operator. This construction ensures:

\begin{itemize}
\item \textbf{Zero-mean property}: $\sum_{i=1}^G \hat{A}_{i,t} = 0$ almost surely for every position $t$, preserving the fixed-point of policy gradient updates.
\item \textbf{Scale invariance}: Microbatch standardization aligns estimator scale across batches, crucial for stable clipping in multilingual settings where reward magnitudes vary substantially.
\item \textbf{Position-aware exploration}: The entropy multiplier $1 + \alpha \sg(H) \gamma^t$ amplifies early token contributions, reflecting their larger causal influence on sequence completion.
\end{itemize}

\subsubsection{Dual Entropy Regularization}

VEPO employs two complementary entropy terms:

\begin{align}
\text{Local entropy} &: \text{within } \hat{A}_{i,t} \text{ for credit assignment modulation} \\
\text{Global entropy} &: \beta \mathbb{E}[H(\pi_\theta(\cdot|q,o_{i,<t}))] \text{ for exploration enforcement}
\end{align}

The local entropy term modulates credit assignment by treating it as a constant weight (stop-gradient), thus avoiding the introduction of complex second-order coupling gradients $\nabla_\theta H$ into the PPO surrogate. The global term explicitly enforces a lower bound on policy entropy through differentiable regularization. This dual structure prevents premature action distribution collapse observed in non-value baselines.

\subsubsection{Temperature-Consistent Importance Sampling}

The importance ratio admits the closed-form expression:

\begin{equation}
r_{i,t}^\tau(\theta) = \exp\left(\frac{\log \pi_\theta(o_{i,t}|q,o_{i,<t}) - \log \pi_{\theta_{\text{old}}}(o_{i,t}|q,o_{i,<t})}{\tau}\right)
\end{equation}

This formulation preserves the martingale property required for on-policy convergence analyses while ensuring consistent temperature scaling across data collection and policy updates.

\subsubsection{Asymmetric Clipping Analysis}

The asymmetric clipping bounds $(1-\epsilon_{\text{low}}, 1+\epsilon_{\text{high}})$ with $\epsilon_{\text{low}} > \epsilon_{\text{high}}$ reflect empirical observations:

\begin{itemize}
\item \textbf{Downward sensitivity}: Ratio decreases (policy becoming less likely to generate observed tokens) correlate strongly with translation quality degradation.
\item \textbf{Upward tolerance}: Moderate ratio increases often correspond to improved translation fluency and adequacy.
\item \textbf{Typical values}: $\epsilon_{\text{low}} = 0.20$, $\epsilon_{\text{high}} = 0.28$ based on extensive hyperparameter search.
\end{itemize}

\subsubsection{Computational Optimizations}

For large vocabularies, entropy computation is approximated by:

\begin{equation}
H(\pi_\theta(\cdot|s)) \approx -\sum_{k \in \text{top-20\%}} \pi_\theta(k|s) \log \pi_\theta(k|s)
\end{equation}

This approximation reduces computational overhead by 80\% while maintaining sufficient accuracy for gradient-based training, as the tail of the probability distribution contributes negligibly to entropy estimates.
\subsection{Experimental Setup Details}
\label{sec:appendix-exp-setup}

\paragraph{Models.} We compare three checkpoints derived from our pipeline---continued pre-training only (\textsc{SFT w/o CPT}), continued pre-training followed by instruction tuning (\textsc{SFT}), and the full VEPO stage---against both multilingual chat models (Gemma3, Qwen2.5/3, Aya, LLaMAX, etc.) and translation-specialized systems (Hunyuan-MT-7B, Qwen-MT-Plus, Seed-X-PPO-7B).

\paragraph{Training budget.} Continued pre-training consumes 200B tokens with adaptive multilingual sampling, while VEPO uses 60k RL steps with per-step batch size of 128 samples. All RL baselines share identical budgets to enable fair comparisons.

\paragraph{Metrics.} Main results report sacreBLEU for three translation groups (xx~$\to$~en, en~$\to$~xx, and xx~$\to$~xx). General capabilities are measured with zero-shot accuracy. Significance is assessed with paired bootstrap resampling across FLORES directions.
\subsection{Detailed BLEU Breakdown}
\label{sec:appendix-blue}

Table~\ref{tab:blue-breakdown} reports the BLEU scores by source and target language pair across all evaluated models, computed with the standard n-gram formulation\citep{papineni-etal-2002-bleu}. Language names from the original dataset are translated into English for readability.

{\scriptsize
\setlength{\tabcolsep}{1.5pt}
\begin{longtable}{@{}llrrrrrrrrrrrrrrrrr@{}}
    \caption{BLEU scores per source--target language pair.}\label{tab:blue-breakdown}\\
    \toprule
    \multirow{2}{*}{\textbf{Source}} & \multirow{2}{*}{\textbf{Target}} & \multicolumn{17}{c}{\textbf{Model BLEU Scores}} \\
    \cmidrule{3-19}
    & & Gem3 & SeedX & QwenMT & Huny & VEPO & Qw2.5 & Qw3 & Apert & Aya & Emma & LLa3.1 & LLaX3 & Mistr & Tower & Google & NLLB \\
    \midrule
    \endfirsthead
    \toprule
    \multirow{2}{*}{\textbf{Source}} & \multirow{2}{*}{\textbf{Target}} & \multicolumn{17}{c}{\textbf{Model BLEU Scores}} \\
    \cmidrule{3-19}
    & & Gem3 & SeedX & QwenMT & Huny & VEPO & Qw2.5 & Qw3 & Apert & Aya & Emma & LLa3.1 & LLaX3 & Mistr & Tower & Google & NLLB \\
    \midrule
    \endhead
    \midrule
    \multicolumn{19}{r}{\emph{Continued on next page}}\\
    \midrule
    \endfoot
    \bottomrule
    \endlastfoot
Thai & Malay & 0.247 & 0.250 & 0.212 & 0.180 & 0.243 & 0.132 & 0.169 & 0.187 & 0.092 & 0.008 & 0.014 & 0.173 & 0.031 & 0.131 & 0.234 & 0.227 \\
Thai & Khmer & 0.078 & 0.006 & 0.051 & 0.090 & 0.091 & 0.018 & 0.042 & 0.005 & 0.002 & 0.077 & 0.012 & 0.063 & 0.002 & 0.003 & 0.095 & 0.072 \\
Thai & Lao & 0.174 & 0.012 & 0.133 & 0.007 & 0.199 & 0.010 & 0.067 & 0.026 & 0.003 & 0.006 & 0.005 & 0.066 & 0.003 & 0.005 & 0.147 & 0.174 \\
Thai & Burmese & 0.102 & 0.002 & 0.135 & 0.121 & 0.161 & 0.013 & 0.048 & 0.095 & 0.001 & 0.067 & 0.015 & 0.047 & 0.003 & 0.002 & 0.116 & 0.066 \\
Thai & Filipino & 0.234 & 0.020 & 0.179 & 0.156 & 0.214 & 0.066 & 0.130 & 0.169 & 0.054 & 0.029 & 0.091 & 0.149 & 0.026 & 0.092 & 0.148 & 0.211 \\
Thai & Indonesian & 0.302 & 0.277 & 0.197 & 0.195 & 0.259 & 0.191 & 0.232 & 0.250 & 0.160 & 0.008 & 0.033 & 0.204 & 0.047 & 0.192 & 0.285 & 0.266 \\
Thai & Vietnamese & 0.318 & 0.321 & 0.240 & 0.234 & 0.296 & 0.212 & 0.268 & 0.278 & 0.190 & 0.015 & 0.030 & 0.256 & 0.056 & 0.210 & 0.321 & 0.294 \\
Thai & English & 0.327 & 0.298 & 0.262 & 0.231 & 0.312 & 0.274 & 0.291 & 0.282 & 0.168 & 0.061 & 0.037 & 0.273 & 0.145 & 0.287 & 0.325 & 0.316 \\
Thai & Chinese & 0.266 & 0.184 & 0.260 & 0.214 & 0.243 & 0.217 & 0.241 & 0.231 & 0.137 & 0.053 & 0.111 & 0.176 & 0.072 & 0.234 & 0.281 & 0.147 \\
Malay & Thai & 0.292 & 0.291 & 0.287 & 0.246 & 0.269 & 0.162 & 0.227 & 0.243 & 0.060 & 0.004 & 0.192 & 0.206 & 0.030 & 0.145 & 0.321 & 0.202 \\
Malay & Khmer & 0.086 & 0.006 & 0.053 & 0.103 & 0.103 & 0.019 & 0.044 & 0.006 & 0.003 & 0.017 & 0.013 & 0.070 & 0.004 & 0.004 & 0.105 & 0.083 \\
Malay & Lao & 0.183 & 0.015 & 0.144 & 0.009 & 0.231 & 0.016 & 0.061 & 0.028 & 0.006 & 0.005 & 0.006 & 0.073 & 0.005 & 0.008 & 0.168 & 0.197 \\
Malay & Burmese & 0.118 & 0.003 & 0.146 & 0.134 & 0.174 & 0.016 & 0.056 & 0.107 & 0.001 & 0.021 & 0.016 & 0.050 & 0.004 & 0.002 & 0.115 & 0.079 \\
Malay & Filipino & 0.259 & 0.027 & 0.229 & 0.208 & 0.273 & 0.079 & 0.159 & 0.203 & 0.099 & 0.028 & 0.177 & 0.207 & 0.050 & 0.125 & 0.048 & 0.274 \\
Malay & Indonesian & 0.385 & 0.374 & 0.366 & 0.270 & 0.312 & 0.239 & 0.307 & 0.263 & 0.336 & 0.080 & 0.254 & 0.254 & 0.113 & 0.239 & 0.381 & 0.368 \\
Malay & Vietnamese & 0.365 & 0.365 & 0.350 & 0.285 & 0.337 & 0.252 & 0.304 & 0.323 & 0.315 & 0.011 & 0.288 & 0.294 & 0.086 & 0.250 & 0.384 & 0.340 \\
Malay & English & 0.440 & 0.438 & 0.441 & 0.341 & 0.435 & 0.373 & 0.397 & 0.413 & 0.379 & 0.079 & 0.363 & 0.402 & 0.274 & 0.421 & 0.504 & 0.466 \\
Malay & Chinese & 0.280 & 0.185 & 0.280 & 0.259 & 0.262 & 0.227 & 0.255 & 0.244 & 0.224 & 0.059 & 0.203 & 0.189 & 0.101 & 0.250 & 0.329 & 0.167 \\
Khmer & Thai & 0.265 & 0.043 & 0.228 & 0.183 & 0.243 & 0.090 & 0.089 & 0.179 & 0.015 & 0.013 & 0.090 & 0.180 & 0.014 & 0.072 & 0.272 & 0.177 \\
Khmer & Malay & 0.237 & 0.054 & 0.200 & 0.155 & 0.234 & 0.070 & 0.141 & 0.171 & 0.032 & 0.093 & 0.005 & 0.160 & 0.017 & 0.072 & 0.245 & 0.239 \\
Khmer & Lao & 0.141 & 0.010 & 0.100 & 0.008 & 0.209 & 0.009 & 0.009 & 0.001 & 0.006 & 0.013 & 0.005 & 0.059 & 0.006 & 0.006 & 0.149 & 0.176 \\
Khmer & Burmese & 0.090 & 0.002 & 0.116 & 0.109 & 0.153 & 0.002 & 0.030 & 0.002 & 0.002 & 0.073 & 0.013 & 0.043 & 0.002 & 0.001 & 0.104 & 0.092 \\
Khmer & Filipino & 0.229 & 0.015 & 0.174 & 0.141 & 0.221 & 0.048 & 0.104 & 0.159 & 0.033 & 0.011 & 0.018 & 0.148 & 0.015 & 0.054 & 0.217 & 0.216 \\
Khmer & Indonesian & 0.279 & 0.062 & 0.238 & 0.168 & 0.260 & 0.096 & 0.183 & 0.223 & 0.044 & 0.082 & 0.008 & 0.180 & 0.022 & 0.095 & 0.291 & 0.266 \\
Khmer & Vietnamese & 0.296 & 0.069 & 0.265 & 0.195 & 0.295 & 0.112 & 0.214 & 0.249 & 0.038 & 0.049 & 0.004 & 0.233 & 0.022 & 0.106 & 0.313 & 0.288 \\
Khmer & English & 0.315 & 0.066 & 0.294 & 0.202 & 0.322 & 0.128 & 0.236 & 0.273 & 0.044 & 0.015 & 0.129 & 0.266 & 0.048 & 0.134 & 0.335 & 0.332 \\
Khmer & Chinese & 0.238 & 0.014 & 0.211 & 0.165 & 0.227 & 0.095 & 0.175 & 0.180 & 0.050 & 0.053 & 0.027 & 0.153 & 0.022 & 0.113 & 0.262 & 0.143 \\
Lao & Thai & 0.297 & 0.016 & 0.281 & 0.038 & 0.275 & 0.085 & 0.134 & 0.093 & 0.016 & 0.041 & 0.110 & 0.201 & 0.007 & 0.092 & 0.316 & 0.215 \\
Lao & Malay & 0.275 & 0.016 & 0.222 & 0.039 & 0.284 & 0.062 & 0.164 & 0.191 & 0.037 & 0.109 & 0.015 & 0.144 & 0.010 & 0.076 & 0.280 & 0.267 \\
Lao & Khmer & 0.087 & 0.002 & 0.042 & 0.041 & 0.109 & 0.009 & 0.011 & 0.002 & 0.004 & 0.053 & 0.009 & 0.065 & 0.005 & 0.004 & 0.103 & 0.084 \\
Lao & Burmese & 0.103 & 0.001 & 0.105 & 0.039 & 0.165 & 0.003 & 0.026 & 0.002 & 0.002 & 0.039 & 0.009 & 0.035 & 0.003 & 0.002 & 0.113 & 0.095 \\
Lao & Filipino & 0.246 & 0.005 & 0.188 & 0.051 & 0.239 & 0.049 & 0.127 & 0.167 & 0.035 & 0.050 & 0.054 & 0.129 & 0.012 & 0.048 & 0.244 & 0.235 \\
Lao & Indonesian & 0.314 & 0.018 & 0.228 & 0.040 & 0.292 & 0.072 & 0.196 & 0.231 & 0.047 & 0.067 & 0.042 & 0.157 & 0.014 & 0.087 & 0.333 & 0.296 \\
Lao & Vietnamese & 0.329 & 0.023 & 0.270 & 0.039 & 0.314 & 0.087 & 0.229 & 0.259 & 0.040 & 0.057 & 0.032 & 0.217 & 0.013 & 0.096 & 0.350 & 0.301 \\
Lao & English & 0.359 & 0.014 & 0.257 & 0.048 & 0.364 & 0.094 & 0.266 & 0.282 & 0.051 & 0.056 & 0.120 & 0.236 & 0.029 & 0.120 & 0.415 & 0.375 \\
Lao & Chinese & 0.247 & 0.005 & 0.235 & 0.050 & 0.231 & 0.072 & 0.184 & 0.189 & 0.054 & 0.055 & 0.030 & 0.126 & 0.013 & 0.102 & 0.295 & 0.150 \\
Burmese & Thai & 0.228 & 0.015 & 0.203 & 0.145 & 0.204 & 0.046 & 0.040 & 0.157 & 0.005 & 0.013 & 0.080 & 0.123 & 0.007 & 0.036 & 0.249 & 0.162 \\
Burmese & Malay & 0.194 & 0.017 & 0.170 & 0.128 & 0.193 & 0.031 & 0.099 & 0.141 & 0.009 & 0.093 & 0.004 & 0.092 & 0.005 & 0.051 & 0.212 & 0.220 \\
Burmese & Khmer & 0.055 & 0.002 & 0.055 & 0.068 & 0.073 & 0.007 & 0.013 & 0.003 & 0.002 & 0.041 & 0.005 & 0.040 & 0.002 & 0.001 & 0.079 & 0.062 \\
Burmese & Lao & 0.113 & 0.004 & 0.092 & 0.004 & 0.169 & 0.003 & 0.015 & 0.010 & 0.003 & 0.023 & 0.002 & 0.039 & 0.001 & 0.001 & 0.129 & 0.149 \\
Burmese & Filipino & 0.200 & 0.006 & 0.155 & 0.124 & 0.188 & 0.031 & 0.077 & 0.140 & 0.013 & 0.084 & 0.055 & 0.096 & 0.009 & 0.039 & 0.197 & 0.205 \\
Burmese & Indonesian & 0.233 & 0.018 & 0.210 & 0.132 & 0.201 & 0.046 & 0.132 & 0.181 & 0.013 & 0.086 & 0.009 & 0.109 & 0.008 & 0.070 & 0.259 & 0.249 \\
Burmese & Vietnamese & 0.260 & 0.020 & 0.235 & 0.166 & 0.254 & 0.057 & 0.164 & 0.218 & 0.014 & 0.057 & 0.020 & 0.173 & 0.010 & 0.077 & 0.286 & 0.269 \\
Burmese & English & 0.263 & 0.020 & 0.254 & 0.173 & 0.268 & 0.061 & 0.181 & 0.229 & 0.016 & 0.046 & 0.153 & 0.198 & 0.023 & 0.104 & 0.317 & 0.304 \\
Burmese & Chinese & 0.207 & 0.004 & 0.191 & 0.140 & 0.189 & 0.049 & 0.132 & 0.163 & 0.013 & 0.326 & 0.063 & 0.124 & 0.010 & 0.091 & 0.244 & 0.131 \\
Filipino & Thai & 0.298 & 0.246 & 0.280 & 0.226 & 0.267 & 0.134 & 0.205 & 0.218 & 0.051 & 0.008 & 0.173 & 0.197 & 0.025 & 0.125 & 0.320 & 0.238 \\
Filipino & Malay & 0.317 & 0.277 & 0.211 & 0.225 & 0.326 & 0.131 & 0.203 & 0.193 & 0.146 & 0.095 & 0.180 & 0.229 & 0.052 & 0.171 & 0.376 & 0.422 \\
Filipino & Khmer & 0.083 & 0.006 & 0.060 & 0.095 & 0.103 & 0.016 & 0.038 & 0.005 & 0.003 & 0.011 & 0.010 & 0.071 & 0.004 & 0.003 & 0.123 & 0.090 \\
Filipino & Lao & 0.164 & 0.015 & 0.136 & 0.008 & 0.227 & 0.013 & 0.046 & 0.023 & 0.005 & 0.558 & 0.006 & 0.064 & 0.004 & 0.005 & 0.197 & 0.232 \\
Filipino & Burmese & 0.122 & 0.003 & 0.128 & 0.132 & 0.168 & 0.015 & 0.050 & 0.093 & 0.002 & 0.019 & 0.017 & 0.050 & 0.004 & 0.002 & 0.128 & 0.122 \\
Filipino & Indonesian & 0.375 & 0.300 & 0.263 & 0.231 & 0.336 & 0.176 & 0.229 & 0.246 & 0.245 & 0.161 & 0.221 & 0.261 & 0.078 & 0.203 & 0.332 & 0.362 \\
Filipino & Vietnamese & 0.368 & 0.315 & 0.139 & 0.264 & 0.347 & 0.194 & 0.279 & 0.294 & 0.254 & 0.056 & 0.268 & 0.285 & 0.067 & 0.204 & 0.499 & 0.467 \\
Filipino & English & 0.481 & 0.394 & 0.454 & 0.328 & 0.454 & 0.299 & 0.382 & 0.395 & 0.307 & 0.535 & 0.399 & 0.403 & 0.239 & 0.382 & 0.457 & 0.416 \\
Filipino & Chinese & 0.287 & 0.129 & 0.271 & 0.237 & 0.267 & 0.191 & 0.237 & 0.236 & 0.175 & 0.029 & 0.185 & 0.184 & 0.084 & 0.235 & 0.396 & 0.205 \\
Indonesian & Thai & 0.304 & 0.304 & 0.299 & 0.256 & 0.276 & 0.172 & 0.237 & 0.254 & 0.064 & 0.254 & 0.200 & 0.209 & 0.030 & 0.161 & 0.283 & 0.171 \\
Indonesian & Malay & 0.332 & 0.351 & 0.313 & 0.268 & 0.299 & 0.220 & 0.262 & 0.240 & 0.163 & 0.037 & 0.222 & 0.215 & 0.083 & 0.228 & 0.245 & 0.218 \\
Indonesian & Khmer & 0.085 & 0.006 & 0.064 & 0.102 & 0.101 & 0.018 & 0.045 & 0.006 & 0.003 & 0.119 & 0.012 & 0.078 & 0.004 & 0.004 & 0.085 & 0.065 \\
Indonesian & Lao & 0.178 & 0.015 & 0.138 & 0.008 & 0.226 & 0.016 & 0.060 & 0.030 & 0.006 & 0.047 & 0.006 & 0.067 & 0.005 & 0.007 & 0.135 & 0.138 \\
Indonesian & Burmese & 0.115 & 0.003 & 0.137 & 0.133 & 0.176 & 0.016 & 0.054 & 0.102 & 0.002 & 0.033 & 0.018 & 0.051 & 0.003 & 0.002 & 0.108 & 0.114 \\
Indonesian & Filipino & 0.262 & 0.025 & 0.236 & 0.209 & 0.273 & 0.085 & 0.170 & 0.211 & 0.101 & 0.081 & 0.181 & 0.211 & 0.049 & 0.124 & 0.214 & 0.189 \\
Indonesian & Vietnamese & 0.371 & 0.377 & 0.362 & 0.296 & 0.349 & 0.270 & 0.321 & 0.341 & 0.350 & 0.491 & 0.310 & 0.307 & 0.087 & 0.270 & 0.299 & 0.241 \\
Indonesian & English & 0.447 & 0.443 & 0.446 & 0.344 & 0.435 & 0.399 & 0.411 & 0.428 & 0.416 & 0.012 & 0.350 & 0.395 & 0.269 & 0.435 & 0.334 & 0.292 \\
Indonesian & Chinese & 0.293 & 0.213 & 0.302 & 0.266 & 0.271 & 0.255 & 0.271 & 0.260 & 0.251 & 0.398 & 0.216 & 0.203 & 0.113 & 0.276 & 0.347 & 0.304 \\
Vietnamese & Thai & 0.280 & 0.252 & 0.277 & 0.232 & 0.254 & 0.155 & 0.222 & 0.237 & 0.065 & 0.285 & 0.184 & 0.200 & 0.029 & 0.149 & 0.300 & 0.198 \\
Vietnamese & Malay & 0.277 & 0.283 & 0.251 & 0.215 & 0.279 & 0.150 & 0.200 & 0.214 & 0.169 & 0.337 & 0.174 & 0.213 & 0.055 & 0.171 & 0.279 & 0.274 \\
Vietnamese & Khmer & 0.079 & 0.006 & 0.052 & 0.099 & 0.093 & 0.018 & 0.043 & 0.005 & 0.003 & 0.365 & 0.013 & 0.065 & 0.003 & 0.004 & 0.096 & 0.078 \\
Vietnamese & Lao & 0.167 & 0.014 & 0.137 & 0.008 & 0.211 & 0.014 & 0.054 & 0.026 & 0.005 & 0.415 & 0.007 & 0.067 & 0.005 & 0.007 & 0.155 & 0.179 \\
Vietnamese & Burmese & 0.112 & 0.003 & 0.141 & 0.130 & 0.165 & 0.014 & 0.050 & 0.099 & 0.002 & 0.334 & 0.015 & 0.051 & 0.004 & 0.002 & 0.110 & 0.074 \\
Vietnamese & Filipino & 0.261 & 0.024 & 0.207 & 0.187 & 0.251 & 0.075 & 0.147 & 0.198 & 0.092 & 0.345 & 0.171 & 0.188 & 0.040 & 0.113 & 0.249 & 0.244 \\
Vietnamese & Indonesian & 0.330 & 0.312 & 0.304 & 0.231 & 0.298 & 0.224 & 0.269 & 0.294 & 0.304 & 0.353 & 0.226 & 0.253 & 0.084 & 0.236 & 0.345 & 0.316 \\
Vietnamese & English & 0.376 & 0.376 & 0.364 & 0.290 & 0.365 & 0.334 & 0.345 & 0.345 & 0.353 & 0.242 & 0.237 & 0.326 & 0.226 & 0.352 & 0.421 & 0.385 \\
Vietnamese & Chinese & 0.273 & 0.218 & 0.283 & 0.247 & 0.252 & 0.235 & 0.250 & 0.244 & 0.242 & 0.290 & 0.189 & 0.183 & 0.105 & 0.253 & 0.316 & 0.164 \\
English & Thai & 0.341 & 0.348 & 0.345 & 0.286 & 0.320 & 0.200 & 0.272 & 0.308 & 0.074 & 0.304 & 0.235 & 0.236 & 0.034 & 0.184 & 0.376 & 0.238 \\
English & Malay & 0.403 & 0.433 & 0.381 & 0.305 & 0.411 & 0.239 & 0.308 & 0.353 & 0.243 & 0.017 & 0.332 & 0.322 & 0.093 & 0.250 & 0.394 & 0.422 \\
English & Khmer & 0.091 & 0.007 & 0.060 & 0.115 & 0.125 & 0.022 & 0.051 & 0.008 & 0.004 & 0.271 & 0.013 & 0.078 & 0.003 & 0.004 & 0.122 & 0.090 \\
English & Lao & 0.205 & 0.017 & 0.172 & 0.010 & 0.274 & 0.017 & 0.075 & 0.034 & 0.006 & 0.242 & 0.006 & 0.074 & 0.005 & 0.007 & 0.197 & 0.232 \\
English & Burmese & 0.129 & 0.003 & 0.166 & 0.153 & 0.208 & 0.019 & 0.066 & 0.126 & 0.002 & 0.183 & 0.020 & 0.062 & 0.004 & 0.002 & 0.128 & 0.122 \\
English & Filipino & 0.354 & 0.031 & 0.297 & 0.250 & 0.352 & 0.120 & 0.220 & 0.294 & 0.132 & 0.350 & 0.272 & 0.270 & 0.069 & 0.177 & 0.332 & 0.362 \\
English & Indonesian & 0.478 & 0.468 & 0.470 & 0.321 & 0.448 & 0.328 & 0.401 & 0.463 & 0.439 & 0.017 & 0.410 & 0.364 & 0.134 & 0.362 & 0.499 & 0.467 \\
English & Vietnamese & 0.421 & 0.455 & 0.414 & 0.339 & 0.415 & 0.316 & 0.380 & 0.416 & 0.406 & 0.017 & 0.377 & 0.350 & 0.108 & 0.311 & 0.457 & 0.416 \\
English & Chinese & 0.341 & 0.271 & 0.355 & 0.327 & 0.329 & 0.307 & 0.328 & 0.316 & 0.304 & 0.017 & 0.271 & 0.242 & 0.183 & 0.350 & 0.396 & 0.205 \\
Chinese & Thai & 0.263 & 0.252 & 0.246 & 0.220 & 0.237 & 0.142 & 0.209 & 0.222 & 0.053 & 0.017 & 0.166 & 0.175 & 0.023 & 0.129 & 0.283 & 0.171 \\
Chinese & Malay & 0.221 & 0.234 & 0.200 & 0.186 & 0.223 & 0.119 & 0.164 & 0.181 & 0.132 & 0.017 & 0.156 & 0.162 & 0.035 & 0.138 & 0.245 & 0.218 \\
Chinese & Khmer & 0.073 & 0.006 & 0.062 & 0.090 & 0.087 & 0.018 & 0.036 & 0.005 & 0.004 & 0.017 & 0.010 & 0.063 & 0.002 & 0.005 & 0.085 & 0.065 \\
Chinese & Lao & 0.135 & 0.011 & 0.112 & 0.007 & 0.174 & 0.010 & 0.043 & 0.020 & 0.004 & 0.017 & 0.004 & 0.042 & 0.003 & 0.005 & 0.135 & 0.138 \\
Chinese & Burmese & 0.097 & 0.002 & 0.133 & 0.122 & 0.164 & 0.014 & 0.055 & 0.101 & 0.001 & 0.017 & 0.016 & 0.048 & 0.002 & 0.002 & 0.107 & 0.114 \\
Chinese & Filipino & 0.212 & 0.019 & 0.171 & 0.160 & 0.198 & 0.058 & 0.120 & 0.160 & 0.067 & 0.017 & 0.134 & 0.146 & 0.029 & 0.088 & 0.214 & 0.189 \\
Chinese & Indonesian & 0.269 & 0.264 & 0.253 & 0.207 & 0.248 & 0.179 & 0.224 & 0.236 & 0.255 & 0.017 & 0.203 & 0.187 & 0.054 & 0.197 & 0.299 & 0.241 \\
Chinese & Vietnamese & 0.306 & 0.319 & 0.289 & 0.258 & 0.292 & 0.181 & 0.274 & 0.276 & 0.288 & 0.017 & 0.248 & 0.247 & 0.060 & 0.222 & 0.334 & 0.292 \\
    Chinese & English & 0.301 & 0.282 & 0.292 & 0.256 & 0.292 & 0.273 & 0.288 & 0.275 & 0.290 & 0.017 & 0.271 & 0.255 & 0.187 & 0.306 & 0.347 & 0.304 \\
\end{longtable}
}

\subsection{Detailed COMET Breakdown}
\label{sec:appendix-comet}

Table~\ref{tab:comet-breakdown} reports COMET scores for every evaluated source--target direction across VEPO and all baselines, complementing the BLEU breakdown in Appendix~\ref{sec:appendix-blue}.

{\scriptsize
\setlength{\tabcolsep}{1.5pt}
\begin{longtable}{@{}llrrrrrrrrrrrrrrrr@{}}
    \caption{COMET scores per source--target language pair.}\label{tab:comet-breakdown}\\
    \toprule
    \multirow{2}{*}{\textbf{Source}} & \multirow{2}{*}{\textbf{Target}} & \multicolumn{16}{c}{\textbf{Model COMET Scores}} \\
    \cmidrule(lr){3-18}
    & & Gem3 & SeedX & QwenMT & Huny & VEPO & Qw2.5 & Qw3 & Apert & Aya & Emma & LLa3.1 & LLaX3 & Mistr & Tower & Google & NLLB \\
    \midrule
    \endfirsthead
    \toprule
    \multirow{2}{*}{\textbf{Source}} & \multirow{2}{*}{\textbf{Target}} & \multicolumn{16}{c}{\textbf{Model COMET Scores}} \\
    \cmidrule(lr){3-18}
    & & Gem3 & SeedX & QwenMT & Huny & VEPO & Qw2.5 & Qw3 & Apert & Aya & Emma & LLa3.1 & LLaX3 & Mistr & Tower & Google & NLLB \\
    \midrule
    \endhead
    \midrule
    \multicolumn{18}{r}{\emph{Continued on next page}}\\
    \midrule
    \endfoot
    \bottomrule
    \endlastfoot
    Thai & Malay & 0.8774 & 0.876 & 0.877 & 0.8729 & 0.8741 & 0.5088 & 0.8537 & 0.8627 & 0.7755 & 0.4776 & 0.7004 & 0.8471 & 0.5788 & 0.7917 & 0.8481 & 0.8572 \\
    Thai & Khmer & 0.7952 & 0.4917 & 0.7807 & 0.8126 & 0.8245 & 0.3895 & 0.6391 & 0.5416 & 0.3694 & 0.7217 & 0.5474 & 0.6985 & 0.4164 & 0.3355 & 0.7682 & 0.7884 \\
    Thai & Lao & 0.8223 & 0.51 & 0.7957 & 0.6894 & 0.8409 & 0.3552 & 0.6412 & 0.5743 & 0.3659 & 0.5655 & 0.5892 & 0.6166 & 0.3353 & 0.4796 & 0.7615 & 0.8295 \\
    Thai & Burmese & 0.8373 & 0.4567 & 0.8487 & 0.8498 & 0.8673 & 0.4155 & 0.6412 & 0.7807 & 0.3781 & 0.7714 & 0.5436 & 0.6405 & 0.388 & 0.3276 & 0.7535 & 0.8205 \\
    Thai & Filipino & 0.8401 & 0.655 & 0.8298 & 0.8244 & 0.8353 & 0.4611 & 0.7514 & 0.8136 & 0.5903 & 0.5316 & 0.7156 & 0.7974 & 0.5538 & 0.6924 & 0.749 & 0.819 \\
    Thai & Indonesian & 0.901 & 0.8988 & 0.8988 & 0.8907 & 0.8898 & 0.5243 & 0.8843 & 0.889 & 0.8209 & 0.4651 & 0.737 & 0.8678 & 0.6252 & 0.8546 & 0.8923 & 0.878 \\
    Thai & Vietnamese & 0.8892 & 0.8867 & 0.8896 & 0.883 & 0.8808 & 0.504 & 0.8762 & 0.8739 & 0.8117 & 0.4553 & 0.7257 & 0.8615 & 0.5719 & 0.8324 & 0.8776 & 0.8685 \\
    Thai & English & 0.8902 & 0.8863 & 0.8899 & 0.8782 & 0.8847 & 0.5233 & 0.8813 & 0.87 & 0.8069 & 0.685 & 0.6816 & 0.8731 & 0.7952 & 0.8778 & 0.8804 & 0.8743 \\
    Thai & Chinese & 0.8794 & 0.8745 & 0.8841 & 0.8778 & 0.8736 & 0.5127 & 0.8727 & 0.8684 & 0.7914 & 0.6146 & 0.8102 & 0.8473 & 0.7292 & 0.8671 & 0.873 & 0.819 \\
    Malay & Thai & 0.8692 & 0.8667 & 0.8694 & 0.8674 & 0.8553 & 0.4569 & 0.8367 & 0.8373 & 0.5409 & 0.5551 & 0.8041 & 0.8153 & 0.4664 & 0.7574 & 0.864 & 0.8242 \\
    Malay & Khmer & 0.7774 & 0.4523 & 0.759 & 0.8048 & 0.8129 & 0.3786 & 0.6218 & 0.5215 & 0.3672 & 0.6755 & 0.5314 & 0.6893 & 0.3848 & 0.3524 & 0.7558 & 0.7828 \\
    Malay & Lao & 0.8097 & 0.4699 & 0.7704 & 0.6461 & 0.8361 & 0.3476 & 0.5987 & 0.5413 & 0.3611 & 0.5203 & 0.3883 & 0.594 & 0.318 & 0.3955 & 0.7544 & 0.818 \\
    Malay & Burmese & 0.8333 & 0.4167 & 0.8383 & 0.849 & 0.8636 & 0.4207 & 0.6512 & 0.7796 & 0.3867 & 0.7784 & 0.542 & 0.6458 & 0.3925 & 0.3405 & 0.7315 & 0.828 \\
    Malay & Filipino & 0.8501 & 0.6247 & 0.8327 & 0.8404 & 0.8459 & 0.4362 & 0.7447 & 0.7946 & 0.6319 & 0.59 & 0.7674 & 0.8042 & 0.5606 & 0.6912 & 0.7353 & 0.8367 \\
    Malay & Indonesian & 0.9181 & 0.9182 & 0.9149 & 0.9138 & 0.9038 & 0.5051 & 0.9016 & 0.895 & 0.9023 & 0.6664 & 0.8935 & 0.8869 & 0.7003 & 0.8818 & 0.915 & 0.9072 \\
    Malay & Vietnamese & 0.8832 & 0.8841 & 0.8817 & 0.8806 & 0.8702 & 0.4708 & 0.8578 & 0.8556 & 0.8606 & 0.5954 & 0.8365 & 0.8432 & 0.5596 & 0.8107 & 0.8834 & 0.8639 \\
    Malay & English & 0.894 & 0.8932 & 0.893 & 0.8842 & 0.8918 & 0.4882 & 0.8794 & 0.8822 & 0.8715 & 0.6742 & 0.866 & 0.8804 & 0.8407 & 0.8778 & 0.8978 & 0.8907 \\
    Malay & Chinese & 0.8639 & 0.8635 & 0.8697 & 0.8708 & 0.8589 & 0.4868 & 0.8545 & 0.8447 & 0.8357 & 0.6041 & 0.8305 & 0.8276 & 0.7354 & 0.8468 & 0.8711 & 0.8098 \\
    Khmer & Thai & 0.8581 & 0.6512 & 0.8556 & 0.8462 & 0.8522 & 0.4609 & 0.7295 & 0.7711 & 0.4606 & 0.6524 & 0.6795 & 0.8166 & 0.4325 & 0.6588 & 0.8496 & 0.8123 \\
    Khmer & Malay & 0.8606 & 0.6687 & 0.857 & 0.846 & 0.8588 & 0.4826 & 0.8141 & 0.8376 & 0.5707 & 0.5084 & 0.4516 & 0.8246 & 0.513 & 0.6724 & 0.8368 & 0.8453 \\
    Khmer & Lao & 0.8063 & 0.4383 & 0.7616 & 0.6677 & 0.8396 & 0.4369 & 0.6654 & 0.4637 & 0.376 & 0.6479 & 0.4812 & 0.6119 & 0.4404 & 0.4293 & 0.754 & 0.8159 \\
    Khmer & Burmese & 0.81 & 0.4041 & 0.8207 & 0.8413 & 0.8579 & 0.4117 & 0.591 & 0.4325 & 0.392 & 0.604 & 0.5353 & 0.63 & 0.3885 & 0.3217 & 0.7387 & 0.8233 \\
    Khmer & Filipino & 0.8345 & 0.5475 & 0.8213 & 0.8137 & 0.8305 & 0.4754 & 0.7228 & 0.8023 & 0.5273 & 0.5005 & 0.4547 & 0.7842 & 0.5054 & 0.619 & 0.817 & 0.8128 \\
    Khmer & Indonesian & 0.8811 & 0.686 & 0.8747 & 0.8624 & 0.8743 & 0.4976 & 0.8353 & 0.8592 & 0.599 & 0.5223 & 0.4594 & 0.8408 & 0.5387 & 0.7165 & 0.8754 & 0.8613 \\
    Khmer & Vietnamese & 0.876 & 0.6836 & 0.8739 & 0.8591 & 0.8698 & 0.485 & 0.8359 & 0.853 & 0.5866 & 0.5455 & 0.4726 & 0.8424 & 0.4999 & 0.706 & 0.8678 & 0.8524 \\
    Khmer & English & 0.8748 & 0.6759 & 0.8714 & 0.8545 & 0.8763 & 0.5095 & 0.8424 & 0.8635 & 0.5871 & 0.4808 & 0.6344 & 0.8564 & 0.6086 & 0.7543 & 0.8669 & 0.8632 \\
    Khmer & Chinese & 0.8582 & 0.6376 & 0.8584 & 0.8474 & 0.8548 & 0.4873 & 0.825 & 0.8387 & 0.5706 & 0.5558 & 0.53 & 0.8174 & 0.5608 & 0.739 & 0.8543 & 0.8093 \\
    Lao & Thai & 0.8686 & 0.5347 & 0.8645 & 0.5859 & 0.8593 & 0.4395 & 0.7585 & 0.6251 & 0.4541 & 0.592 & 0.6764 & 0.8107 & 0.4399 & 0.6397 & 0.8663 & 0.8308 \\
    Lao & Malay & 0.8685 & 0.5192 & 0.8598 & 0.5998 & 0.867 & 0.4651 & 0.8148 & 0.8345 & 0.5581 & 0.4646 & 0.4544 & 0.8046 & 0.4622 & 0.6311 & 0.8527 & 0.853 \\
    Lao & Khmer & 0.7796 & 0.357 & 0.7379 & 0.5851 & 0.8223 & 0.4228 & 0.6395 & 0.4829 & 0.3845 & 0.6404 & 0.5064 & 0.6895 & 0.4269 & 0.3454 & 0.7699 & 0.7867 \\
    Lao & Burmese & 0.8237 & 0.3396 & 0.7731 & 0.6355 & 0.8592 & 0.3896 & 0.5821 & 0.4228 & 0.3871 & 0.7017 & 0.5037 & 0.6057 & 0.3894 & 0.3252 & 0.7462 & 0.821 \\
    Lao & Filipino & 0.8403 & 0.4352 & 0.8244 & 0.6208 & 0.8343 & 0.4693 & 0.7274 & 0.7944 & 0.5273 & 0.4442 & 0.5359 & 0.7672 & 0.4763 & 0.5864 & 0.8034 & 0.8186 \\
    Lao & Indonesian & 0.8862 & 0.5362 & 0.8772 & 0.608 & 0.8785 & 0.4753 & 0.8358 & 0.8525 & 0.5865 & 0.4798 & 0.5206 & 0.817 & 0.4876 & 0.6709 & 0.8887 & 0.8663 \\
    Lao & Vietnamese & 0.8777 & 0.5401 & 0.8708 & 0.5909 & 0.8698 & 0.4635 & 0.8301 & 0.8424 & 0.571 & 0.4911 & 0.5183 & 0.8188 & 0.4642 & 0.6514 & 0.8786 & 0.8557 \\
    Lao & English & 0.8793 & 0.5099 & 0.8737 & 0.5908 & 0.8807 & 0.4873 & 0.8407 & 0.8553 & 0.5803 & 0.5519 & 0.6795 & 0.8317 & 0.5094 & 0.6973 & 0.8858 & 0.8703 \\
    Lao & Chinese & 0.8613 & 0.5356 & 0.8574 & 0.5858 & 0.8576 & 0.4659 & 0.8196 & 0.8275 & 0.5594 & 0.5716 & 0.5444 & 0.7892 & 0.512 & 0.6763 & 0.8632 & 0.8073 \\
    Burmese & Thai & 0.8511 & 0.5809 & 0.8523 & 0.8312 & 0.841 & 0.4595 & 0.6714 & 0.7998 & 0.43 & 0.6321 & 0.6905 & 0.7883 & 0.4255 & 0.6075 & 0.8452 & 0.8112 \\
    Burmese & Malay & 0.8529 & 0.5939 & 0.8516 & 0.8359 & 0.8425 & 0.4844 & 0.7907 & 0.8261 & 0.5439 & 0.5019 & 0.4221 & 0.7981 & 0.4912 & 0.6526 & 0.8337 & 0.854 \\
    Burmese & Khmer & 0.7468 & 0.4073 & 0.7621 & 0.786 & 0.8054 & 0.4037 & 0.5781 & 0.5215 & 0.3822 & 0.6282 & 0.5061 & 0.6611 & 0.4895 & 0.3253 & 0.753 & 0.7675 \\
    Burmese & Lao & 0.7794 & 0.4128 & 0.7622 & 0.6343 & 0.8204 & 0.4153 & 0.5798 & 0.5099 & 0.3728 & 0.5964 & 0.38 & 0.5671 & 0.2983 & 0.3732 & 0.7487 & 0.8024 \\
    Burmese & Filipino & 0.8318 & 0.513 & 0.8213 & 0.8111 & 0.8257 & 0.4853 & 0.7015 & 0.7943 & 0.5018 & 0.5176 & 0.5707 & 0.7573 & 0.5029 & 0.5997 & 0.805 & 0.8233 \\
    Burmese & Indonesian & 0.8727 & 0.6055 & 0.8715 & 0.8512 & 0.8636 & 0.4964 & 0.8172 & 0.847 & 0.5667 & 0.527 & 0.4432 & 0.8176 & 0.5192 & 0.6989 & 0.8724 & 0.8694 \\
    Burmese & Vietnamese & 0.8649 & 0.5947 & 0.8647 & 0.8437 & 0.8558 & 0.4813 & 0.8058 & 0.8341 & 0.5538 & 0.5142 & 0.4647 & 0.8222 & 0.4872 & 0.6751 & 0.8621 & 0.8582 \\
    Burmese & English & 0.8687 & 0.6009 & 0.871 & 0.8483 & 0.869 & 0.5134 & 0.8284 & 0.8573 & 0.5546 & 0.4874 & 0.7366 & 0.8461 & 0.5704 & 0.7379 & 0.8723 & 0.8729 \\
    Burmese & Chinese & 0.8519 & 0.5612 & 0.8541 & 0.835 & 0.8433 & 0.4791 & 0.8073 & 0.8281 & 0.5319 & 0.5455 & 0.631 & 0.8085 & 0.5252 & 0.7137 & 0.8519 & 0.8147 \\
    Filipino & Thai & 0.8605 & 0.8382 & 0.8561 & 0.8524 & 0.8456 & 0.4445 & 0.8054 & 0.8014 & 0.5221 & 0.5238 & 0.7758 & 0.8047 & 0.4483 & 0.7238 & 0.8522 & 0.8155 \\
    Filipino & Malay & 0.8777 & 0.8614 & 0.8617 & 0.8691 & 0.8722 & 0.458 & 0.7999 & 0.8118 & 0.7906 & 0.5144 & 0.8009 & 0.837 & 0.5611 & 0.7679 & 0.8551 & 0.8657 \\
    Filipino & Khmer & 0.7747 & 0.4293 & 0.7503 & 0.7947 & 0.8045 & 0.3853 & 0.5913 & 0.4993 & 0.3611 & 0.5651 & 0.5112 & 0.6668 & 0.3642 & 0.3355 & 0.748 & 0.776 \\
    Filipino & Lao & 0.8021 & 0.4473 & 0.7667 & 0.6351 & 0.8275 & 0.3457 & 0.5616 & 0.5115 & 0.3566 & 0.4764 & 0.3677 & 0.5847 & 0.2998 & 0.3739 & 0.7417 & 0.8063 \\
    Filipino & Burmese & 0.8343 & 0.3965 & 0.8318 & 0.8455 & 0.8591 & 0.4183 & 0.6254 & 0.7175 & 0.3772 & 0.758 & 0.5324 & 0.6382 & 0.3842 & 0.3343 & 0.7275 & 0.8202 \\
    Filipino & Indonesian & 0.8988 & 0.8808 & 0.8832 & 0.8865 & 0.8871 & 0.4716 & 0.8227 & 0.8358 & 0.8294 & 0.5367 & 0.8238 & 0.8565 & 0.6171 & 0.8012 & 0.9005 & 0.8816 \\
    Filipino & Vietnamese & 0.872 & 0.8498 & 0.8662 & 0.86 & 0.8572 & 0.4492 & 0.8184 & 0.8266 & 0.8044 & 0.4842 & 0.8029 & 0.8262 & 0.528 & 0.7597 & 0.8697 & 0.849 \\
    Filipino & English & 0.8841 & 0.8599 & 0.8753 & 0.8659 & 0.878 & 0.4702 & 0.8509 & 0.8583 & 0.8147 & 0.5923 & 0.8516 & 0.8613 & 0.792 & 0.8423 & 0.8888 & 0.8752 \\
    Filipino & Chinese & 0.8558 & 0.8225 & 0.8552 & 0.8536 & 0.8453 & 0.4706 & 0.8271 & 0.8231 & 0.7776 & 0.532 & 0.7996 & 0.809 & 0.6868 & 0.8132 & 0.8609 & 0.7949 \\
    Indonesian & Thai & 0.8749 & 0.8739 & 0.8777 & 0.8738 & 0.8644 & 0.4638 & 0.8522 & 0.8514 & 0.5462 & 0.5688 & 0.8158 & 0.825 & 0.4658 & 0.7675 & 0.8674 & 0.8328 \\
    Indonesian & Malay & 0.9045 & 0.9063 & 0.9055 & 0.903 & 0.9035 & 0.5139 & 0.8968 & 0.897 & 0.8722 & 0.6736 & 0.8954 & 0.8838 & 0.6599 & 0.8671 & 0.8872 & 0.9005 \\
    Indonesian & Khmer & 0.7823 & 0.4582 & 0.7613 & 0.8062 & 0.8151 & 0.382 & 0.6241 & 0.5282 & 0.365 & 0.6913 & 0.5312 & 0.6892 & 0.3886 & 0.3423 & 0.7537 & 0.7866 \\
    Indonesian & Lao & 0.8096 & 0.4755 & 0.7655 & 0.6464 & 0.8374 & 0.3468 & 0.5954 & 0.55 & 0.3618 & 0.5377 & 0.3846 & 0.589 & 0.3173 & 0.3909 & 0.7447 & 0.8172 \\
    Indonesian & Burmese & 0.8313 & 0.4219 & 0.8374 & 0.8508 & 0.865 & 0.4237 & 0.652 & 0.7858 & 0.3862 & 0.7847 & 0.5436 & 0.653 & 0.3897 & 0.337 & 0.7263 & 0.8284 \\
    Indonesian & Filipino & 0.853 & 0.6329 & 0.8365 & 0.8409 & 0.848 & 0.4428 & 0.7503 & 0.8096 & 0.6463 & 0.6079 & 0.7784 & 0.8132 & 0.5626 & 0.6966 & 0.8227 & 0.8412 \\
    Indonesian & Vietnamese & 0.8903 & 0.8889 & 0.8913 & 0.8875 & 0.8795 & 0.4782 & 0.8714 & 0.8732 & 0.8814 & 0.6108 & 0.8558 & 0.8539 & 0.5656 & 0.8315 & 0.8998 & 0.8856 \\
    Indonesian & English & 0.8997 & 0.8989 & 0.9002 & 0.8895 & 0.8966 & 0.4962 & 0.8912 & 0.8932 & 0.893 & 0.6712 & 0.8736 & 0.8852 & 0.8473 & 0.8896 & 0.8943 & 0.8738 \\
    Indonesian & Chinese & 0.8736 & 0.8725 & 0.8797 & 0.8765 & 0.8665 & 0.4948 & 0.8672 & 0.86 & 0.8607 & 0.6167 & 0.8441 & 0.841 & 0.7497 & 0.8632 & 0.8688 & 0.8263 \\
    Vietnamese & Thai & 0.8737 & 0.8633 & 0.8771 & 0.8708 & 0.8643 & 0.4677 & 0.8532 & 0.8519 & 0.5504 & 0.5145 & 0.8187 & 0.8313 & 0.4651 & 0.7701 & 0.8877 & 0.8426 \\
    Vietnamese & Malay & 0.8756 & 0.8758 & 0.8744 & 0.8729 & 0.8731 & 0.4887 & 0.8526 & 0.8542 & 0.8467 & 0.4586 & 0.8305 & 0.8506 & 0.6019 & 0.7902 & 0.8673 & 0.814 \\
    Vietnamese & Khmer & 0.7882 & 0.4521 & 0.7735 & 0.8122 & 0.8251 & 0.3805 & 0.6416 & 0.5395 & 0.37 & 0.7 & 0.5408 & 0.6845 & 0.3931 & 0.344 & 0.7593 & 0.7669 \\
    Vietnamese & Lao & 0.8138 & 0.4696 & 0.7771 & 0.654 & 0.8418 & 0.3454 & 0.6021 & 0.5602 & 0.3608 & 0.5385 & 0.3875 & 0.6026 & 0.3374 & 0.3908 & 0.7413 & 0.7968 \\
    Vietnamese & Burmese & 0.8338 & 0.4176 & 0.8424 & 0.8484 & 0.8636 & 0.4213 & 0.6532 & 0.7743 & 0.3811 & 0.7603 & 0.5443 & 0.6409 & 0.3883 & 0.33 & 0.7321 & 0.8241 \\
    Vietnamese & Filipino & 0.8316 & 0.621 & 0.8129 & 0.8161 & 0.8256 & 0.4394 & 0.7284 & 0.7966 & 0.6192 & 0.5686 & 0.7565 & 0.7892 & 0.54 & 0.6711 & 0.7917 & 0.8034 \\
    Vietnamese & Indonesian & 0.9005 & 0.8999 & 0.8985 & 0.894 & 0.8918 & 0.5098 & 0.8854 & 0.8887 & 0.8934 & 0.5278 & 0.8641 & 0.8754 & 0.6656 & 0.8556 & 0.8943 & 0.8738 \\
    Vietnamese & English & 0.8812 & 0.8817 & 0.8813 & 0.8729 & 0.8776 & 0.4936 & 0.8745 & 0.8746 & 0.8754 & 0.6089 & 0.8307 & 0.8676 & 0.8268 & 0.8725 & 0.8881 & 0.8725 \\
    Vietnamese & Chinese & 0.8774 & 0.8781 & 0.8821 & 0.88 & 0.8709 & 0.5051 & 0.8729 & 0.8669 & 0.8691 & 0.6133 & 0.8495 & 0.8485 & 0.7636 & 0.869 & 0.882 & 0.8654 \\
    English & Thai & 0.8951 & 0.8937 & 0.8979 & 0.8935 & 0.887 & 0.4773 & 0.8748 & 0.8746 & 0.5566 & 0.6488 & 0.8488 & 0.8462 & 0.4679 & 0.7985 & 0.8673 & 0.8140 \\
    English & Malay & 0.9022 & 0.9051 & 0.9027 & 0.9015 & 0.9018 & 0.4878 & 0.8802 & 0.8894 & 0.8728 & 0.7139 & 0.8822 & 0.8695 & 0.6266 & 0.807 & 0.8444 & 0.8533 \\
    English & Khmer & 0.8107 & 0.4166 & 0.7876 & 0.8359 & 0.8473 & 0.3943 & 0.6634 & 0.5549 & 0.3766 & 0.771 & 0.557 & 0.7119 & 0.4127 & 0.3625 & 0.7593 & 0.7669 \\
    English & Lao & 0.8339 & 0.4387 & 0.7963 & 0.6894 & 0.8657 & 0.3527 & 0.637 & 0.5758 & 0.3669 & 0.6124 & 0.4244 & 0.6003 & 0.3552 & 0.4038 & 0.7413 & 0.7968 \\
    English & Burmese & 0.862 & 0.3756 & 0.8692 & 0.8779 & 0.8928 & 0.4439 & 0.6922 & 0.8214 & 0.398 & 0.8273 & 0.5743 & 0.6986 & 0.4256 & 0.3551 & 0.7321 & 0.8241 \\
    English & Filipino & 0.8611 & 0.6245 & 0.8431 & 0.8478 & 0.8526 & 0.4375 & 0.7625 & 0.8268 & 0.6438 & 0.5863 & 0.8007 & 0.8105 & 0.5618 & 0.7025 & 0.7917 & 0.8034 \\
    English & Indonesian & 0.9254 & 0.9236 & 0.9256 & 0.9192 & 0.9152 & 0.5106 & 0.9092 & 0.9139 & 0.9172 & 0.6491 & 0.9056 & 0.8883 & 0.6879 & 0.8826 & 0.8943 & 0.8738 \\
    English & Vietnamese & 0.9008 & 0.903 & 0.903 & 0.9 & 0.8912 & 0.4867 & 0.8895 & 0.8886 & 0.8942 & 0.6617 & 0.8785 & 0.8631 & 0.588 & 0.841 & 0.8881 & 0.8725 \\
    English & Chinese & 0.891 & 0.8921 & 0.897 & 0.8987 & 0.8867 & 0.5055 & 0.8877 & 0.8772 & 0.88 & 0.6424 & 0.8713 & 0.8568 & 0.8237 & 0.8906 & 0.882 & 0.8654 \\
    Chinese & Thai & 0.8748 & 0.8723 & 0.8753 & 0.8743 & 0.8671 & 0.4729 & 0.8543 & 0.8545 & 0.5446 & 0.5942 & 0.8168 & 0.8283 & 0.4602 & 0.773 & 0.8751 & 0.8610 \\
    Chinese & Malay & 0.8679 & 0.8704 & 0.8673 & 0.8688 & 0.8687 & 0.4923 & 0.8496 & 0.8558 & 0.8394 & 0.5408 & 0.841 & 0.8381 & 0.5623 & 0.7818 & 0.8709 & 0.8722 \\
    Chinese & Khmer & 0.787 & 0.4623 & 0.7701 & 0.8078 & 0.8251 & 0.3858 & 0.6328 & 0.5361 & 0.3679 & 0.7537 & 0.5263 & 0.6647 & 0.388 & 0.3429 & 0.8751 & 0.8731 \\
    Chinese & Lao & 0.8054 & 0.4784 & 0.7673 & 0.6488 & 0.8351 & 0.3485 & 0.5973 & 0.5417 & 0.3599 & 0.591 & 0.3756 & 0.5629 & 0.3116 & 0.3719 & 0.8787 & 0.8749 \\
    Chinese & Burmese & 0.8293 & 0.4281 & 0.847 & 0.8489 & 0.8667 & 0.4207 & 0.6734 & 0.7982 & 0.3822 & 0.7868 & 0.543 & 0.6467 & 0.3966 & 0.3376 & 0.8779 & 0.8745 \\
    Chinese & Filipino & 0.8235 & 0.6283 & 0.8105 & 0.8142 & 0.8196 & 0.4451 & 0.7311 & 0.7953 & 0.6251 & 0.5728 & 0.762 & 0.7821 & 0.5481 & 0.6699 & 0.8709 & 0.8722 \\
    Chinese & Indonesian & 0.8918 & 0.8928 & 0.8933 & 0.8899 & 0.8862 & 0.5143 & 0.8821 & 0.883 & 0.8859 & 0.569 & 0.8731 & 0.8626 & 0.6312 & 0.8519 & 0.8584 & 0.8660 \\
    Chinese & Vietnamese & 0.8883 & 0.8883 & 0.8898 & 0.8878 & 0.8833 & 0.49 & 0.8801 & 0.879 & 0.8825 & 0.4904 & 0.8658 & 0.8585 & 0.5601 & 0.8391 & 0.861 & 0.8454 \\
    Chinese & English & 0.8749 & 0.8745 & 0.8779 & 0.8731 & 0.8751 & 0.5027 & 0.8738 & 0.8709 & 0.8722 & 0.6584 & 0.866 & 0.861 & 0.8454 & 0.8772 & 0.8738 & 0.8772 \\
\end{longtable}
}
\normalsize

\FloatBarrier
\subsection{Detailed chrF Breakdown}
\label{sec:appendix-chrf}

Table~\ref{tab:chrf-breakdown} reports chrF scores for every evaluated source--target direction across VEPO and all baselines, complementing the BLEU and COMET breakdowns in Appendices~\ref{sec:appendix-blue} and~\ref{sec:appendix-comet}.

{\scriptsize
\setlength{\tabcolsep}{1.5pt}
\begin{longtable}{@{}llrrrrrrrrrrrrrrrr@{}}
    \caption{chrF scores per source--target language pair.}\label{tab:chrf-breakdown}\\
    \toprule
    \multirow{2}{*}{\textbf{Source}} & \multirow{2}{*}{\textbf{Target}} & \multicolumn{16}{c}{\textbf{Model chrF Scores}} \\
    \cmidrule(lr){3-18}
    & & Gem3 & SeedX & QwenMT & Huny & VEPO & Qw2.5 & Qw3 & Apert & Aya & Emma & LLa3.1 & LLaX3 & Mistr & Tower & Google & NLLB \\
    \midrule
    \endfirsthead
    \toprule
    \multirow{2}{*}{\textbf{Source}} & \multirow{2}{*}{\textbf{Target}} & \multicolumn{16}{c}{\textbf{Model chrF Scores}} \\
    \cmidrule(lr){3-18}
    & & Gem3 & SeedX & QwenMT & Huny & VEPO & Qw2.5 & Qw3 & Apert & Aya & Emma & LLa3.1 & LLaX3 & Mistr & Tower & Google & NLLB \\
    \midrule
    \endhead
    \midrule
    \multicolumn{18}{r}{\emph{Continued on next page}}\\
    \midrule
    \endfoot
    \bottomrule
    \endlastfoot
    Thai & Malay & 58.31 & 57.57 & 55.44 & 55.74 & 57.95 & 23.09 & 51.79 & 53.82 & 42.32 & 6.43 & 4.89 & 50.21 & 35.48 & 48.86 & 58.05 & 57.71 \\
    Thai & Khmer & 36.93 & 2.85 & 34.34 & 39.40 & 42.28 & 13.25 & 28.35 & 15.82 & 11.60 & 35.03 & 18.45 & 30.47 & 1.76 & 11.74 & 41.98 & 41.85 \\
    Thai & Lao & 45.85 & 4.08 & 41.23 & 2.48 & 48.60 & 11.28 & 31.41 & 18.32 & 10.51 & 2.03 & 4.79 & 27.54 & 3.21 & 9.20 & 46.82 & 48.23 \\
    Thai & Burmese & 39.40 & 0.71 & 42.25 & 44.15 & 46.55 & 17.47 & 29.83 & 36.38 & 10.15 & 31.80 & 16.20 & 25.48 & 15.08 & 11.84 & 44.23 & 44.18 \\
    Thai & Filipino & 54.12 & 18.07 & 49.96 & 51.67 & 53.14 & 23.36 & 44.68 & 49.49 & 36.54 & 14.57 & 31.06 & 46.40 & 32.88 & 41.20 & 48.17 & 53.69 \\
    Thai & Indonesian & 60.90 & 58.45 & 58.20 & 56.20 & 57.78 & 21.94 & 55.93 & 57.70 & 48.26 & 9.50 & 10.19 & 52.13 & 37.91 & 52.99 & 60.14 & 58.28 \\
    Thai & Vietnamese & 52.90 & 53.38 & 51.56 & 49.98 & 51.14 & 14.68 & 48.79 & 50.00 & 41.03 & 9.84 & 5.96 & 46.57 & 28.68 & 44.09 & 54.21 & 52.74 \\
    Thai & English & 61.02 & 60.64 & 60.07 & 57.14 & 59.96 & 19.80 & 58.72 & 58.80 & 46.20 & 23.22 & 9.04 & 55.28 & 44.48 & 58.76 & 61.19 & 60.53 \\
    Thai & Chinese & 31.87 & 27.60 & 31.70 & 30.94 & 31.01 & 3.20 & 29.94 & 29.69 & 20.08 & 8.31 & 15.89 & 24.94 & 15.53 & 29.27 & 31.84 & 28.46 \\
    Malay & Thai & 53.36 & 52.86 & 52.85 & 52.08 & 51.18 & 13.34 & 47.37 & 48.52 & 26.65 & 11.77 & 43.31 & 44.29 & 23.83 & 40.03 & 57.23 & 50.36 \\
    Malay & Khmer & 36.97 & 3.09 & 34.63 & 40.44 & 43.76 & 12.33 & 28.31 & 16.32 & 11.94 & 31.98 & 18.65 & 31.46 & 2.42 & 12.51 & 44.05 & 43.62 \\
    Malay & Lao & 45.17 & 4.43 & 40.82 & 3.34 & 50.31 & 10.65 & 28.23 & 17.24 & 11.31 & 3.19 & 10.82 & 26.76 & 4.77 & 10.64 & 49.35 & 50.17 \\
    Malay & Burmese & 39.99 & 0.74 & 43.32 & 44.90 & 47.39 & 17.23 & 30.36 & 36.98 & 9.68 & 35.66 & 16.96 & 26.26 & 15.65 & 11.78 & 44.11 & 46.74 \\
    Malay & Filipino & 55.42 & 18.43 & 52.66 & 54.52 & 56.27 & 21.98 & 45.69 & 49.79 & 39.83 & 21.55 & 47.66 & 50.01 & 36.99 & 42.75 & 48.36 & 57.09 \\
    Malay & Indonesian & 65.20 & 65.01 & 63.80 & 61.35 & 60.58 & 21.69 & 59.59 & 56.78 & 61.51 & 25.02 & 56.17 & 56.29 & 48.39 & 55.05 & 63.15 & 62.29 \\
    Malay & Vietnamese & 55.33 & 56.22 & 54.47 & 52.57 & 53.08 & 14.26 & 50.20 & 51.44 & 50.92 & 16.92 & 48.43 & 49.16 & 32.45 & 45.34 & 59.37 & 57.40 \\
    Malay & English & 68.09 & 68.67 & 67.91 & 63.93 & 68.02 & 19.21 & 64.47 & 66.24 & 62.65 & 25.37 & 59.14 & 64.38 & 57.52 & 65.84 & 73.87 & 70.94 \\
    Malay & Chinese & 33.12 & 29.00 & 31.53 & 34.08 & 32.77 & 3.04 & 31.07 & 30.44 & 28.42 & 7.76 & 27.53 & 26.77 & 19.15 & 30.84 & 35.90 & 28.89 \\
    Khmer & Thai & 50.66 & 26.66 & 49.50 & 47.13 & 49.32 & 14.24 & 21.20 & 39.45 & 17.69 & 1.22 & 23.59 & 42.75 & 19.35 & 31.55 & 52.70 & 48.06 \\
    Khmer & Malay & 57.18 & 34.42 & 54.93 & 52.70 & 56.98 & 22.29 & 48.81 & 51.73 & 29.39 & 3.02 & 1.86 & 48.78 & 28.70 & 38.75 & 57.92 & 57.29 \\
    Khmer & Lao & 42.79 & 3.45 & 35.02 & 3.11 & 49.58 & 0.30 & 3.27 & 1.35 & 11.38 & 2.73 & 4.51 & 25.89 & 2.50 & 7.73 & 47.45 & 49.78 \\
    Khmer & Burmese & 35.89 & 0.59 & 40.14 & 42.71 & 46.24 & 0.10 & 17.48 & 2.31 & 9.73 & 0.53 & 14.70 & 24.77 & 5.36 & 10.78 & 43.80 & 44.55 \\
    Khmer & Filipino & 53.59 & 16.98 & 50.15 & 49.66 & 53.51 & 23.45 & 41.58 & 48.40 & 30.93 & 3.06 & 5.70 & 45.27 & 27.92 & 36.05 & 51.17 & 53.41 \\
    Khmer & Indonesian & 59.00 & 34.15 & 56.89 & 52.88 & 57.64 & 22.09 & 51.10 & 55.20 & 29.48 & 3.07 & 2.44 & 49.52 & 29.73 & 41.20 & 60.16 & 58.68 \\
    Khmer & Vietnamese & 51.15 & 27.95 & 50.08 & 46.26 & 50.61 & 14.58 & 43.98 & 47.69 & 22.36 & 3.32 & 2.25 & 44.83 & 22.08 & 33.82 & 54.71 & 53.76 \\
    Khmer & English & 59.29 & 32.50 & 58.49 & 53.24 & 60.18 & 19.57 & 52.82 & 57.40 & 26.13 & 4.26 & 27.97 & 53.64 & 28.51 & 42.12 & 61.45 & 61.57 \\
    Khmer & Chinese & 30.00 & 6.96 & 29.13 & 27.15 & 30.13 & 3.02 & 24.54 & 26.76 & 9.19 & 4.46 & 4.92 & 22.90 & 9.16 & 18.29 & 31.36 & 27.63 \\
    Lao & Thai & 53.51 & 19.50 & 51.94 & 21.41 & 52.14 & 13.95 & 27.72 & 22.10 & 16.61 & 2.26 & 31.72 & 43.82 & 11.28 & 33.47 & 56.23 & 51.67 \\
    Lao & Malay & 59.08 & 22.21 & 54.99 & 28.71 & 59.59 & 21.85 & 49.94 & 52.21 & 28.23 & 3.88 & 4.12 & 46.79 & 23.95 & 36.79 & 59.77 & 59.54 \\
    Lao & Khmer & 36.54 & 2.06 & 28.59 & 20.48 & 44.29 & 1.35 & 4.05 & 4.33 & 11.75 & 32.23 & 13.79 & 31.01 & 1.89 & 9.41 & 43.58 & 44.48 \\
    Lao & Burmese & 38.58 & 0.51 & 34.24 & 23.11 & 46.71 & 2.22 & 14.68 & 1.83 & 9.87 & 28.56 & 12.80 & 23.15 & 7.06 & 10.08 & 44.46 & 45.97 \\
    Lao & Filipino & 54.40 & 13.62 & 49.73 & 30.03 & 53.79 & 22.26 & 42.94 & 47.37 & 30.29 & 4.31 & 20.57 & 43.26 & 24.82 & 33.13 & 54.08 & 55.44 \\
    Lao & Indonesian & 60.41 & 22.94 & 57.05 & 28.00 & 58.86 & 21.38 & 51.30 & 54.34 & 28.56 & 5.33 & 13.34 & 46.90 & 25.10 & 38.26 & 63.01 & 60.97 \\
    Lao & Vietnamese & 52.59 & 19.44 & 49.90 & 21.93 & 51.63 & 14.10 & 44.69 & 46.63 & 21.58 & 6.29 & 9.14 & 42.32 & 18.66 & 31.00 & 55.83 & 54.82 \\
    Lao & English & 61.81 & 19.43 & 59.31 & 26.61 & 62.85 & 19.32 & 54.39 & 56.53 & 25.76 & 15.91 & 34.17 & 50.04 & 19.60 & 37.94 & 65.50 & 65.03 \\
    Lao & Chinese & 30.81 & 3.73 & 29.58 & 10.58 & 30.84 & 2.76 & 24.66 & 25.99 & 8.66 & 9.57 & 6.24 & 20.34 & 6.67 & 15.40 & 33.00 & 28.06 \\
    Burmese & Thai & 47.16 & 21.14 & 46.17 & 42.37 & 44.92 & 13.89 & 11.29 & 38.64 & 9.29 & 0.34 & 27.42 & 37.41 & 16.76 & 21.56 & 49.26 & 45.76 \\
    Burmese & Malay & 53.29 & 27.57 & 50.86 & 49.19 & 50.46 & 21.48 & 42.35 & 47.57 & 26.60 & 1.49 & 1.13 & 42.24 & 24.73 & 34.03 & 52.57 & 54.14 \\
    Burmese & Khmer & 30.46 & 1.24 & 32.70 & 34.89 & 38.73 & 9.58 & 13.20 & 13.03 & 9.35 & 11.91 & 15.19 & 26.75 & 0.58 & 8.38 & 41.06 & 40.91 \\
    Burmese & Lao & 37.69 & 1.81 & 35.18 & 2.17 & 43.53 & 1.36 & 11.39 & 12.73 & 9.45 & 0.71 & 8.93 & 22.23 & 1.86 & 4.54 & 43.39 & 45.51 \\
    Burmese & Filipino & 50.70 & 15.55 & 46.93 & 46.97 & 49.76 & 23.68 & 36.07 & 44.74 & 28.13 & 1.92 & 22.04 & 40.05 & 26.22 & 31.39 & 49.78 & 52.23 \\
    Burmese & Indonesian & 54.89 & 27.10 & 53.13 & 49.02 & 52.79 & 21.34 & 44.79 & 50.34 & 26.33 & 1.88 & 2.67 & 43.39 & 25.99 & 36.04 & 57.36 & 57.19 \\
    Burmese & Vietnamese & 47.50 & 20.18 & 46.06 & 42.42 & 46.05 & 14.01 & 36.95 & 43.05 & 18.94 & 2.79 & 6.41 & 39.11 & 18.46 & 28.17 & 51.63 & 51.12 \\
    Burmese & English & 54.77 & 24.86 & 54.24 & 50.03 & 54.89 & 19.04 & 46.01 & 52.76 & 23.31 & 4.48 & 37.96 & 47.85 & 22.74 & 36.26 & 59.78 & 59.61 \\
    Burmese & Chinese & 26.61 & 3.54 & 25.57 & 23.23 & 25.38 & 2.57 & 19.35 & 22.88 & 5.15 & 1.37 & 9.58 & 19.89 & 5.35 & 14.19 & 28.61 & 25.45 \\
    Filipino & Thai & 53.40 & 49.08 & 51.53 & 50.86 & 50.61 & 13.56 & 45.07 & 44.51 & 24.99 & 5.19 & 41.13 & 43.46 & 22.99 & 38.66 & 55.72 & 50.79 \\
    Filipino & Malay & 61.72 & 57.88 & 55.41 & 58.17 & 61.58 & 21.31 & 51.29 & 50.00 & 46.32 & 13.44 & 48.53 & 53.37 & 38.92 & 49.63 & 63.77 & 63.76 \\
    Filipino & Khmer & 37.65 & 3.02 & 34.33 & 39.91 & 43.48 & 13.11 & 26.19 & 14.84 & 11.55 & 23.84 & 17.57 & 30.43 & 2.35 & 11.78 & 43.59 & 42.98 \\
    Filipino & Lao & 44.13 & 4.30 & 40.05 & 3.36 & 49.64 & 10.73 & 26.07 & 14.26 & 11.04 & 2.88 & 10.25 & 25.94 & 3.51 & 10.24 & 49.72 & 51.63 \\
    Filipino & Burmese & 41.00 & 0.72 & 42.80 & 44.86 & 47.67 & 17.72 & 29.42 & 32.35 & 9.97 & 35.23 & 16.86 & 25.91 & 16.33 & 12.08 & 44.17 & 45.98 \\
    Filipino & Indonesian & 64.77 & 58.88 & 57.86 & 58.64 & 62.25 & 21.08 & 53.45 & 53.66 & 53.46 & 13.68 & 51.19 & 55.47 & 41.96 & 50.88 & 62.33 & 63.02 \\
    Filipino & Vietnamese & 55.96 & 52.48 & 54.48 & 51.84 & 53.71 & 14.38 & 47.83 & 49.48 & 46.47 & 8.51 & 45.87 & 48.03 & 30.29 & 40.98 & 69.46 & 67.49 \\
    Filipino & English & 69.83 & 65.23 & 67.07 & 63.32 & 68.54 & 19.50 & 63.28 & 64.96 & 56.53 & 19.86 & 62.92 & 63.68 & 53.91 & 62.69 & 68.49 & 67.70 \\
    Filipino & Chinese & 33.77 & 24.61 & 32.59 & 32.88 & 32.44 & 3.11 & 29.12 & 29.54 & 23.52 & 5.27 & 25.40 & 25.91 & 17.26 & 28.74 & 38.52 & 30.26 \\
    Indonesian & Thai & 54.04 & 53.80 & 53.81 & 52.75 & 51.67 & 13.47 & 48.42 & 49.46 & 27.09 & 15.13 & 44.18 & 45.05 & 23.74 & 41.73 & 54.29 & 50.39 \\
    Indonesian & Malay & 63.05 & 64.02 & 61.33 & 61.08 & 59.99 & 21.63 & 57.28 & 55.40 & 51.87 & 26.08 & 53.86 & 54.05 & 45.13 & 54.42 & 59.28 & 59.85 \\
    Indonesian & Khmer & 38.03 & 3.06 & 34.98 & 40.62 & 43.86 & 12.51 & 28.70 & 16.49 & 11.96 & 33.56 & 18.54 & 31.44 & 2.38 & 12.10 & 43.71 & 43.86 \\
    Indonesian & Lao & 44.92 & 4.37 & 40.34 & 3.34 & 49.88 & 10.84 & 28.57 & 18.12 & 11.40 & 3.56 & 10.70 & 25.95 & 4.91 & 10.58 & 48.43 & 49.68 \\
    Indonesian & Burmese & 39.74 & 0.74 & 43.01 & 44.92 & 47.61 & 17.74 & 30.47 & 36.80 & 9.74 & 35.53 & 17.05 & 26.71 & 15.57 & 11.73 & 43.90 & 46.35 \\
    Indonesian & Filipino & 55.76 & 18.46 & 53.84 & 54.60 & 56.55 & 22.48 & 46.61 & 51.09 & 40.23 & 22.55 & 48.60 & 50.58 & 37.14 & 42.64 & 54.69 & 55.27 \\
    Indonesian & Vietnamese & 56.09 & 57.30 & 55.88 & 53.69 & 53.99 & 14.13 & 51.83 & 53.26 & 54.24 & 17.63 & 50.68 & 49.94 & 32.68 & 47.29 & 58.64 & 57.87 \\
    Indonesian & English & 68.76 & 69.61 & 68.91 & 64.53 & 68.30 & 19.29 & 66.24 & 67.76 & 66.56 & 23.37 & 58.49 & 64.47 & 58.53 & 67.42 & 69.08 & 68.05 \\
    Indonesian & Chinese & 34.31 & 30.61 & 35.10 & 34.69 & 33.35 & 3.17 & 32.39 & 32.25 & 31.22 & 8.05 & 28.97 & 27.71 & 19.90 & 33.13 & 36.69 & 32.47 \\
    Vietnamese & Thai & 52.08 & 51.11 & 51.82 & 50.84 & 49.69 & 13.41 & 46.88 & 47.69 & 26.84 & 7.92 & 42.06 & 44.06 & 23.52 & 40.36 & 55.90 & 53.40 \\
    Vietnamese & Malay & 58.92 & 59.02 & 57.46 & 57.08 & 58.73 & 21.97 & 53.08 & 54.03 & 49.44 & 12.70 & 48.65 & 52.94 & 39.31 & 50.69 & 58.59 & 54.58 \\
    Vietnamese & Khmer & 36.87 & 2.98 & 34.46 & 39.68 & 42.72 & 12.91 & 28.42 & 16.05 & 11.93 & 32.52 & 18.62 & 29.75 & 2.93 & 11.56 & 42.94 & 42.02 \\
    Vietnamese & Lao & 43.71 & 4.31 & 39.96 & 3.28 & 48.54 & 11.03 & 27.71 & 17.35 & 11.17 & 3.78 & 11.07 & 27.03 & 6.03 & 10.78 & 48.12 & 48.98 \\
    Vietnamese & Burmese & 39.81 & 0.72 & 43.29 & 44.72 & 46.84 & 17.83 & 30.05 & 36.46 & 9.88 & 33.16 & 16.80 & 25.80 & 16.36 & 11.48 & 43.78 & 45.22 \\
    Vietnamese & Filipino & 55.26 & 18.16 & 51.78 & 52.85 & 54.25 & 22.62 & 45.02 & 50.23 & 39.67 & 19.97 & 47.67 & 48.34 & 35.28 & 41.45 & 51.89 & 52.75 \\
    Vietnamese & Indonesian & 61.64 & 60.49 & 60.78 & 58.06 & 59.32 & 21.56 & 57.20 & 59.28 & 59.57 & 13.06 & 52.69 & 55.07 & 43.01 & 54.59 & 61.96 & 59.98 \\
    Vietnamese & English & 63.04 & 64.11 & 62.96 & 59.94 & 62.62 & 19.55 & 61.34 & 61.74 & 61.13 & 18.28 & 47.38 & 58.74 & 51.99 & 61.49 & 66.42 & 64.50 \\
    Vietnamese & Chinese & 32.58 & 30.33 & 33.39 & 33.04 & 31.93 & 3.21 & 30.84 & 30.32 & 30.43 & 7.67 & 26.61 & 26.31 & 19.65 & 31.17 & 34.86 & 29.54 \\
    English & Thai & 57.35 & 58.17 & 57.66 & 55.98 & 55.67 & 13.53 & 51.56 & 54.39 & 28.15 & 26.87 & 47.61 & 47.63 & 24.60 & 44.77 & 61.71 & 56.87 \\
    English & Malay & 67.96 & 70.11 & 66.79 & 64.40 & 68.27 & 21.84 & 61.20 & 64.03 & 55.77 & 38.52 & 61.91 & 61.31 & 45.59 & 57.47 & 65.27 & 64.87 \\
    English & Khmer & 39.25 & 3.17 & 36.00 & 42.44 & 47.24 & 13.15 & 30.25 & 17.79 & 11.99 & 42.76 & 19.01 & 32.37 & 2.25 & 12.65 & 47.22 & 47.37 \\
    English & Lao & 47.45 & 4.52 & 43.97 & 3.48 & 54.81 & 11.35 & 30.98 & 19.74 & 11.37 & 5.37 & 9.94 & 26.79 & 4.64 & 10.88 & 53.14 & 55.14 \\
    English & Burmese & 41.62 & 0.77 & 46.01 & 47.04 & 50.92 & 18.73 & 32.46 & 39.95 & 9.93 & 37.92 & 17.86 & 28.75 & 12.89 & 12.35 & 47.75 & 49.78 \\
    English & Filipino & 62.69 & 18.89 & 58.03 & 58.48 & 61.49 & 22.15 & 50.48 & 57.25 & 42.64 & 24.45 & 55.21 & 54.44 & 40.19 & 46.58 & 61.76 & 63.47 \\
    English & Indonesian & 72.11 & 71.95 & 71.63 & 65.97 & 69.35 & 21.48 & 66.43 & 70.45 & 69.16 & 27.21 & 66.70 & 62.73 & 49.80 & 64.17 & 72.87 & 71.32 \\
    English & Vietnamese & 60.22 & 63.73 & 60.76 & 57.44 & 59.39 & 14.77 & 56.80 & 59.15 & 58.75 & 25.15 & 56.24 & 53.56 & 35.58 & 50.93 & 67.26 & 65.16 \\
    English & Chinese & 38.76 & 35.69 & 40.14 & 40.23 & 38.92 & 3.27 & 37.68 & 36.97 & 36.09 & 16.85 & 33.68 & 31.56 & 26.24 & 39.93 & 41.98 & 35.57 \\
    Chinese & Thai & 50.65 & 50.32 & 50.31 & 50.21 & 49.10 & 13.79 & 46.00 & 47.08 & 25.68 & 11.30 & 40.72 & 42.39 & 23.35 & 39.82 & 53.57 & 53.82 \\
    Chinese & Malay & 56.00 & 56.44 & 54.78 & 55.34 & 55.66 & 21.94 & 50.57 & 52.21 & 46.90 & 15.25 & 49.21 & 49.16 & 37.00 & 48.52 & 56.57 & 57.11 \\
    Chinese & Khmer & 35.26 & 2.79 & 34.32 & 38.86 & 41.11 & 13.39 & 27.16 & 15.29 & 11.82 & 36.97 & 17.32 & 27.37 & 1.85 & 12.06 & 42.07 & 42.62 \\
    Chinese & Lao & 40.85 & 3.98 & 37.65 & 2.75 & 45.54 & 11.64 & 26.61 & 16.13 & 10.79 & 2.59 & 10.47 & 22.86 & 3.22 & 10.74 & 46.14 & 47.79 \\
    Chinese & Burmese & 37.95 & 0.65 & 43.24 & 44.16 & 46.27 & 18.41 & 31.30 & 37.31 & 9.82 & 32.06 & 16.70 & 25.56 & 16.92 & 12.17 & 43.61 & 46.88 \\
    Chinese & Filipino & 52.46 & 17.92 & 49.42 & 51.45 & 51.48 & 22.91 & 43.23 & 48.04 & 38.08 & 19.01 & 45.73 & 45.35 & 34.67 & 40.30 & 53.40 & 54.19 \\
    Chinese & Indonesian & 58.66 & 58.23 & 58.42 & 56.95 & 56.85 & 21.71 & 54.98 & 56.36 & 56.91 & 16.83 & 52.87 & 51.12 & 40.11 & 52.95 & 58.87 & 58.69 \\
    Chinese & Vietnamese & 51.93 & 53.63 & 51.93 & 51.14 & 50.58 & 14.12 & 49.15 & 49.83 & 50.40 & 10.73 & 46.70 & 45.82 & 30.03 & 44.79 & 54.92 & 53.85 \\
    Chinese & English & 59.24 & 59.47 & 59.95 & 58.50 & 59.34 & 19.42 & 58.82 & 58.41 & 58.19 & 25.99 & 56.75 & 54.82 & 51.87 & 60.06 & 60.59 & 61.49 \\
\end{longtable}
}
\normalsize

\FloatBarrier
\subsection{Case Study: Translation Behavior Analysis}
\label{sec:appendix-demo-case}

To illustrate the diversity of translation behaviors observed during data curation, Table~\ref{tab:demo-multi-assistant} presents a concrete example where the same Vietnamese idiom is translated by four different assistant configurations with varying decoding strategies.

\begin{table}[!htb]
    \centering
    \small
    \setlength{\tabcolsep}{6pt}
    \begin{tabular}{p{0.16\linewidth}p{0.78\linewidth}}
        \toprule
        \textbf{Role} & \textbf{Utterance} \\
        \midrule
        System & You are a helpful assistant. \\
        & \rule{0pt}{0.8em}\\
        User & \makecell[l]{Please translate the following sentence into English: \textit{Cho ngap phai ruoi} } \\
        & \rule{0pt}{0.8em}\\
        AssistantA & By sheer chance. \\
        & \rule{0pt}{0.8em}\\
        AssistantB & A dog yawns and catches a fly. \\
        & \rule{0pt}{0.8em}\\
        AssistantC & The dog yawned and hit a fly (Literal meaning: when the dog yawns, it bumps into a fly.)\\
        & This is a metaphor indicating that while doing something you intended to avoid or miss someone or something, you unexpectedly ran into him or it. In other words, you tried to steer clear but ended up encountering it anyway.\\
        & Comparable English idioms include "things go contrary to one's wishes" or "the more you try to hide it, the more obvious it becomes," both signalling intention-outcome mismatch. \\
        & \rule{0pt}{0.8em}\\
        AssistantD & The dog hit a fly, the dog hit a fly---yes, it keeps hitting the same fly. That is literally what it means, it just keeps saying the dog is hitting the fly again and again. (Repetitive and partially incorrect.) \\
        \bottomrule
    \end{tabular}
    \caption{Example conversation showing four different translation approaches: idiomatic, literal, explanatory, and noisy outputs in response to the same Vietnamese idiom.}
    \label{tab:demo-multi-assistant}
\end{table}

\subsubsection{Translation Quality Patterns}

Table~\ref{tab:demo-multi-assistant} illustrates four distinct translation approaches, revealing key patterns in model behavior:

\textbf{Optimal Translation (AssistantA):} The response "By sheer chance" captures the idiomatic meaning effectively while maintaining brevity. This represents the ideal translation—accurate and succinct without unnecessary elaboration.

\textbf{Literal Translation (AssistantB):} "A dog yawns and catches a fly" preserves the original structure but fails to convey the cultural meaning of the idiom, potentially confusing readers unfamiliar with Vietnamese expressions.

\textbf{Over-explanation (AssistantC):} This response exemplifies the overgeneration problem frequently associated with high entropy outputs. Despite being informative, the 72-word response significantly diverges from the core translation objective. The model's uncertainty manifests as excessive elaboration.

\textbf{Degraded Output (AssistantD):} Features inaccuracies, repetition, and self-dialogue patterns. This output typifies problems associated with excessive response length and high entropy, where the model becomes trapped in loops of self-correction and redundant explanation.

\subsubsection{Entropy-Length Correlation}

Our analysis reveals a strong correlation between response length and translation quality. Optimal translations efficiently convey meaning without superfluous context or explanation. Longer responses often coincide with increased token entropy, indicating model uncertainty that manifests as verbose outputs.

These patterns align with our broader findings that entropy loss and response length serve as key diagnostic indicators for translation quality. When models generate unnecessarily long responses, they typically exhibit higher entropy across generated tokens—a sign of uncertainty that degrades translation performance. VEPO addresses this by monitoring and constraining these metrics, guiding models toward concise, accurate translations like AssistantA rather than verbose or literal approaches.
\subsection{KL Divergence Approximation Analysis}
\label{sec:kl-analysis}

In our entropy diagnostics, we employ three Monte-Carlo estimators---k1, k2, and k3---to approximate KL divergence between policy and reference distributions. Each estimator offers a distinct bias-variance profile when computed from sampled translations.

\subsubsection{Method Comparison}

\textbf{k1 Method (Naive Estimator):} Directly uses the negative expectation of log ratios:
\begin{equation}
\text{k1} = -\mathbb{E}_{x \sim q} \left[ \log \frac{p(x)}{q(x)} \right]
\end{equation}
This estimator is unbiased but exhibits high variance and may yield negative values on small batches even though KL divergence is non-negative.

\textbf{k2 Method (Squared Log Ratio):} Approximates KL divergence using the squared log ratio:
\begin{equation}
\text{k2} = \frac{1}{2} \mathbb{E}_{x \sim q} \left[ \left( \log \frac{p(x)}{q(x)} \right)^2 \right]
\end{equation}
It introduces bias yet typically maintains lower variance. k2 corresponds to an f-divergence whose second-order expansion matches KL divergence when $p$ is close to $q$.

\textbf{k3 Method (Bregman Divergence):} Evaluates the expectation of an exponential transform:
\begin{equation}
\text{k3} = \mathbb{E}_{x \sim q} \left[ \frac{p(x)}{q(x)} - 1 - \log \frac{p(x)}{q(x)} \right]
\end{equation}
This estimator remains unbiased with comparatively low variance, measuring the vertical gap between $\log(x)$ and its tangent approximation.

\subsubsection{Practical Selection}

Choosing among these estimators depends on the desired bias-variance trade-off:
\begin{itemize}
    \item \textbf{k1}: Unbiased but high variance; sensitive to entropy spikes.
    \item \textbf{k2}: Slightly biased yet low variance; reliable near convergence.
    \item \textbf{k3}: Unbiased with low variance; preferred default in our monitoring stack.
\end{itemize}

For VEPO's training diagnostics, k2 and k3 deliver smoother curves than k1, particularly under entropy collapse scenarios that arise during reinforcement learning. These approximation methods build on established Monte Carlo techniques\citep{joschu2020klapprox}. Their stability provides more trustworthy signals when tuning entropy regularization and reward scaling.
\subsection{Language Model Assistance in Paper Preparation}
\label{sec:appendix-llm-assistance}

In the preparation of this manuscript, we utilized Large Language Models (LLMs) to assist with language polishing and writing optimization, following emerging practices in academic writing assistance. Specifically, LLMs were employed for:

\begin{itemize}
    \item \textbf{Language fluency improvement}: Enhancing sentence structure and expression to improve readability and adherence to academic writing standards
    \item \textbf{Grammar and spelling verification}: Identifying and correcting potential grammatical errors and spelling issues
    \item \textbf{Terminology consistency}: Ensuring consistent usage of technical terms throughout the manuscript
    \item \textbf{Clarity enhancement}: Improving the articulation of complex concepts to make them more accessible and comprehensible
\end{itemize}

This section contains additional information and supplementary materials.
It is important to emphasize that all core technical content, experimental design, data analysis, and scientific conclusions represent original work by the authors. LLMs were used solely as language polishing tools and did not participate in any substantive research content creation or formation of academic viewpoints. All technical contributions and innovations in this research stem entirely from the independent research work of the author team.

We believe that the reasonable use of advanced language technology tools to enhance academic writing quality, while maintaining academic integrity, represents beneficial practice that facilitates better communication of research findings and promotes scholarly exchange.

\end{document}